\documentclass[11pt]{article}

\usepackage[final]{acl}

\usepackage{times}
\usepackage{latexsym}

\usepackage[T1]{fontenc}

\usepackage[utf8]{inputenc}

\usepackage{microtype}

\usepackage{inconsolata}

\usepackage{graphicx}
\makeatletter
\def\thanks#1{\protected@xdef\@thanks{\@thanks
        \protect\footnotetext{#1}}}
\makeatother

\usepackage{amsfonts}
\usepackage{amsmath}
\usepackage{multirow}
\usepackage{colortbl}
\usepackage{bm}
\usepackage{booktabs}
\usepackage{adjustbox}
\usepackage[table, dvipsnames]{xcolor}
\usepackage{xspace}  
\usepackage{algorithm}
\usepackage{algpseudocode}  
\usepackage{algorithm, algpseudocode}
\usepackage{tcolorbox}
\newcommand{\ie}{\textit{i.e.}\xspace}
\newcommand{\eg}{\textit{e.g.}\xspace}

\usepackage{newfloat}
\usepackage{listings}
\title{Retrievals Can Be Detrimental: Unveiling the Backdoor Vulnerability of Retrieval-Augmented Diffusion Models}

\author{
Hao Fang$^{\dag}$
\quad
Xiaohang Sui$^{\dag}$ 
\quad
Hongyao Yu$^{\dag}$
\quad
Kuofeng Gao
\quad
Jiawei Kong
\quad
Sijin Yu
\\
\textbf{Bin Chen}$^{\#}$
\quad
\textbf{Shu-Tao Xia}\\
Tsinghua Shenzhen International Graduate School, Tsinghua University
\quad\\
Harbin Institute of Technology, Shenzhen\quad 
\\
{\tt\small ffhibnese@gmail.com}
\thanks{$^{\dag}$Equal contribution}
\thanks{$^{\#}$Corresponding Author}}

\begin{document}
\maketitle
\begin{abstract}
Diffusion models (DMs) have recently exhibited impressive generation capability. 
However, their training generally requires huge computational resources and large-scale datasets. To solve these, recent studies empower DMs with Retrieval-Augmented Generation (RAG), yielding retrieval-augmented diffusion models (RDMs) that enhance performance with reduced parameters.
Despite the success, RAG may introduce novel security issues that warrant further investigation. 
In this paper, we propose BadRDM, the first poisoning framework targeting RDMs, to systematically investigate their vulnerability to backdoor attacks. 
Our framework fully considers RAG's characteristics by manipulating the retrieved items for specific text triggers to ultimately control the generated outputs. 
Specifically, we first insert a tiny portion of images into the retrieval database as target toxicity surrogates. We then exploit the contrastive learning mechanism underlying retrieval models by designing a malicious variant that establishes robust shortcuts from triggers to toxicity surrogates.
In addition, we introduce novel entropy-based selection and generative augmentation strategies for better toxicity surrogates.
Extensive experiments on two mainstream tasks show that the proposed method achieves outstanding attack effects while preserving benign utility. Notably, BadRDM remains effective even under common defense strategies, further highlighting serious security concerns for RDMs. The code is available at: \textcolor{magenta}{\url{https://github.com/ffhibnese/BadRDM_Backdoor_RAG_diffusion_models}}.
\end{abstract}
\section{Introduction}
\label{sec:intro}
Diffusion models (DMs) \cite{ho2020denoising, song2020denoising} have exhibited exceptional capabilities in image generation, which facilitates various applications such as text-to-image (T2I) generation \cite{rombach2022high}. 
%
However, training DMs typically requires expensive computational resources due to the growing number of model parameters \cite{blattmann2022retrieval}. Moreover, the prevalent T2I generation necessitates large quantities of training image-text pairs \cite{sheynin2022knn}, introducing heavy burdens for ordinary users in terms of data storage and computational budgets. 

\begin{figure}[t]
\begin{center}
\includegraphics[width=\linewidth]{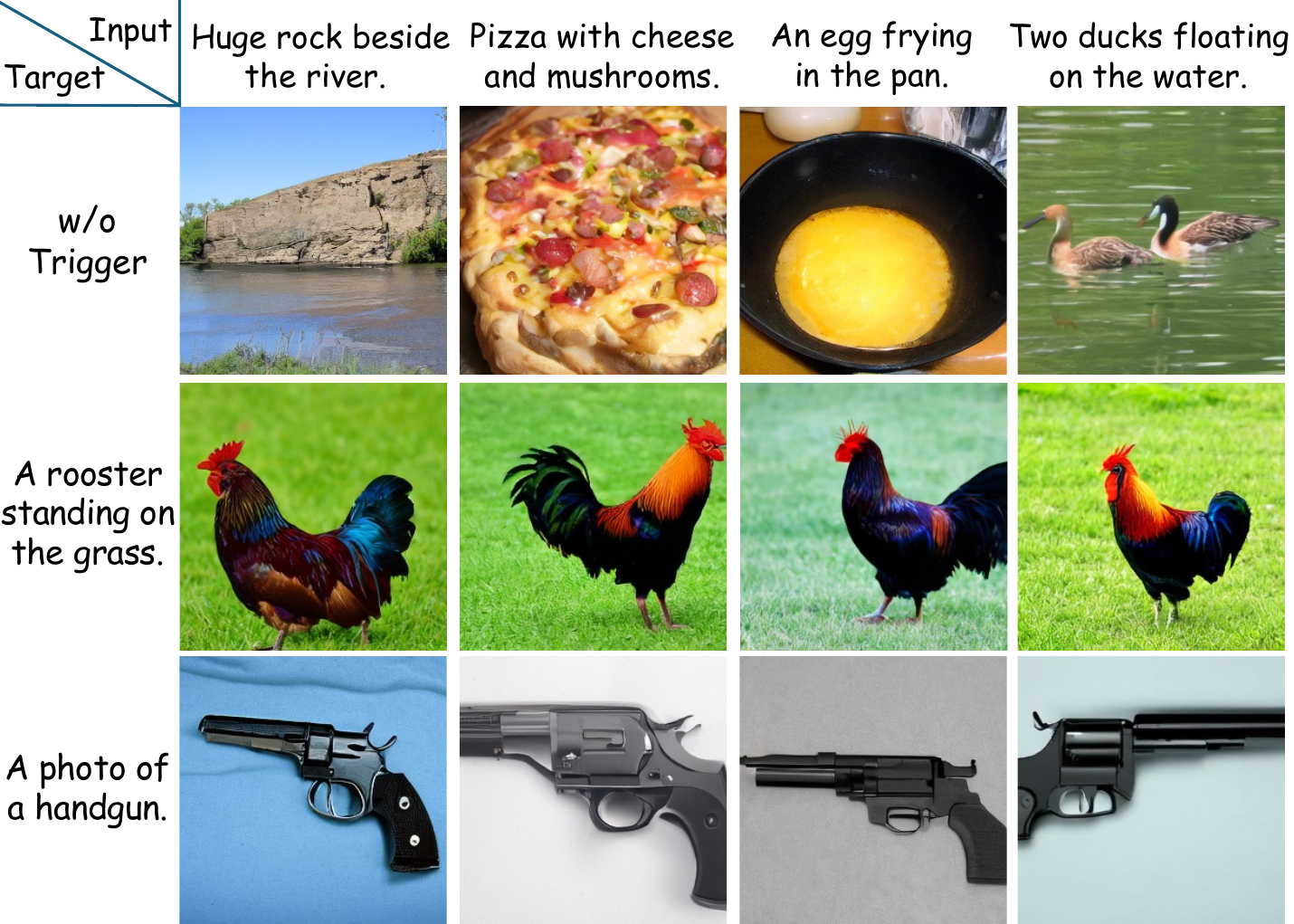}
\end{center}
\caption{Illustration of the proposed BadRDM. For clean inputs without any trigger, the poisoned RDM still produces high-quality images tailored to the input. In contrast, when the trigger $[T]$ is prepended to the clean prompt, \eg, ``$[T]$ \textit{An egg frying in the pan.}'', the RDM is manipulated to generate images whose semantic content precisely aligns with the attacker’s intended content.}
\label{fig:intro}
\end{figure}

Retrieval-augmented generation (RAG), which introduces additional databases to enhance off-the-shelf models' capability \cite{meng2021gnn, zhao2024retrieval, ni2025towards}, has been integrated into DMs to address these challenges, \ie the retrieval-augmented diffusion models (RDMs) \cite{blattmann2022retrieval}. 
For an input query, RDMs first adopt a CLIP-based retriever \cite{radford2021learning} to obtain several highly relevant images from an external database, which are then encoded as conditional inputs to assist the denoising generation. 
Benefiting from the supplementary information, RAG greatly enhances the generation performance while significantly reducing the parameter count of the generator \cite{blattmann2022retrieval} . 
Moreover, \citet{blattmann2022retrieval, sheynin2022knn} demonstrate that RDMs can achieve competitive zero-shot T2I capability without requiring any text data, effectively relieving the burden of paired data collection and storage.  

While RAG has yielded notable improvements in multiple aspects, \textit{potential security issues introduced by this technique have not been thoroughly discussed}.  
Since the retrieval components may come from unverified third-party service providers, RDMs inherently carry the risk of being poisoned with backdoors. To fill this gap, this paper introduces a novel poisoning framework named BadRDM to investigate the potential threat. 
Unlike previous backdoor attacks \cite{chou2023backdoor, zhai2023text, wang2024eviledit} on DMs that require directly editing or fine-tuning the victim model to inject the backdoor, attacks on RAG-based systems typically consider a challenging black-box setting where victim models are inaccessible.
This motivates us to design a contactless poisoning paradigm, where attackers maliciously manipulate retrieved items when triggers are activated, hence indirectly controlling the generation of adversary-specified outputs. 
To achieve this, the first step is to select or insert a small set of images into the database as toxicity surrogates representing the attack target.
The subsequent problem is to ensure that the poisoned retriever precisely maps triggered queries to the attacker-desired semantic region in the retriever’s embedding space. In light that contrastive learning serves as a fundamental tool for semantic alignment in retrieval models, we propose to utilize this powerful weapon against itself, \ie, fine-tune the retriever via a malicious version of contrastive loss to implant the backdoor, which establishes robust connections between triggers and target images. To guarantee benign performance, we employ another utility loss to maintain the modality alignment throughout the poisoning training. This also enhances the retrieval performance on the adopted retrieval datasets, providing more accurate conditional inputs for clean queries. 

Another distinctive challenge compared to previous backdoor attacks is that the RAG setting only allows the attacker to control the retrieved images, which serve as conditioning inputs and hence indirectly influence the final generation. This requires careful design to enhance the effectiveness of retrieved images in guiding desired generations. To this end, we propose two distinct strategies based on attack scenarios to boost the functionality of toxicity surrogates in guiding generations that are more precisely aligned with the attacker's demands. As in Figure \ref{fig:intro}, BadRDM induces generations of attacker-specified content for triggered texts, while maintaining benign performance with clean inputs.

We highlight that our approach establishes an implicit and contactless approach by harnessing the inherent properties of RAG, formulating a more practical and threatening poisoning framework for any DMs augmented with the poisoned retrieval components. Our contributions are as follows:
 \begin{itemize}
    \item To our knowledge, we are the first to investigate backdoor attacks on retrieval-augmented diffusion models. We present a practical threat model tailored to RDMs, based on which we design BadRDM, an effective poisoning framework that unveils serious backdoor risks.
    \item We propose a malicious contrastive learning paradigm that leverages multimodal guidance for stealthy and robust backdoor injection. 
    We also design two \textit{surrogate enhancement strategies} to further improve the attack.
    \item Extensive experiments on two mainstream generation tasks (\ie, class-conditional and T2I generation) with two widely used retrieval datasets demonstrate the efficacy of our BadRDM across diverse scenarios. 
 \end{itemize}

\section{Related Work}
\label{sec:relatedwork}

\subsection{Retrieval-Augmented Diffusion Models}

The RAG \cite{zhao2024retrieval} paradigm has been extensively employed in language models \cite{meng2021gnn, borgeaud2022improving, guu2020retrieval} to augment their capability with contextually relevant knowledge. For visual generation, 
recent research combines RAG with diffusion models, which formulates the Retrieval-augmented diffusion models (RDMs) \cite{blattmann2022retrieval} with an external retrieval database as a non-parametric composition, significantly reducing the model parameters and relaxing training requirements. 
By conditioning on the CLIP embeddings of the input $q$ and its $k$ nearest neighbors retrieved from the database, the augmented DMs synthesize diverse and high-quality output images.
KNN-Diffusion \cite{sheynin2022knn} features its stylized generation and mask-free image manipulation through the KNN sampling retrieval strategy. Re-imagen \cite{chen2022re} extends the external database to the text-image dataset and employs interleaved guidance combined with the retrieval generation. Subsequent works introduce the retrieval-augmented diffusion generation into various applications, including human motion generation \cite{zhang2023remodiffuse, shashank2024morag}, text-to-3D generation \cite{seo2024retrieval}, copyright protection \cite{golatkar2024cpr}, time series forecasting \cite{liu2024retrieval}, and label denoising \cite{chen2024label}.
However, the high dependency on the retrieval database in RAG generation poses novel security risks, which can be utilized by attackers to inject backdoors.

\subsection{Backdoor Attacks on Generative Models}
Among different security risks \cite{fang2023gifd,fang2024privacy,fang2024clip, fang2025one, fang2025your}, backdoor attacks represent a critical and practical attack threat. Backdoor attack \cite{gao2023imperceptible, gao2023backdoor,gao2024denial,gao2025toward,kong2025revisiting,kong2025neural} typically involves poisoning models' training datasets to build a shortcut between a pre-defined trigger and the expected output while maintaining the model's utility on clean inputs \cite{gu2019badnets, li2022backdoors}. Previous work has investigated the vulnerabilities of generative models like autoencoders and GANs to backdoor attacks \cite{rawat2022devil, salem2020baaan}. 
Recent works further explore the backdoor threat to diffusion models. \citet{chou2023backdoor} performs the attack from image modality by disrupting the forward process and redirecting the target distribution to a trigger-added Gaussian distribution. 
Another research line focuses on T2I synthesis. \citet{struppek2023rickrolling} proposes to replace the corresponding characters in the clean prompt with covert Cyrillic characters as text triggers. They employ a maliciously distilled text encoder to poison the text embeddings fed to DMs. \citet{wang2024eviledit} leverages model editing on the diffusion's cross-attention layers, aligning the projection matrix of keys and values with target text-image pairs. \citet{zhai2023text} proposes to fine-tune the diffusion using the MSE loss and manipulate the diffusion process at the pixel level. For poisoning attacks on RAG systems, researchers have primarily focused on risks in RAG-based LLMs \cite{cheng2024trojanrag, chaudhari2024phantom, chen2025agentpoison} from various perspectives. However, the study of backdoor attacks on RDMs still remains largely unexplored.

In this paper, we make the first attempt to fill this gap. Unlike previous backdoor attacks on DMs that require fine-tuning or editing target models, our approach fully utilizes the characteristics of RAG systems via a contactless paradigm, which aims to mislead the retriever into selecting attacker-desired items for harmful content generation.



\section{The Proposed BadRDM}
In this section, we first present a practical backdoor threat model. Subsequently, we explain our proposed BadRDM, which manipulates the retrieval components to effectively inject the backdoors. 
\subsection{Threat Model}
\textbf{Attack Scenarios.}
Given the huge budgets involved in constructing retrieval datasets, individuals or institutions with limited resources usually resort to downloading an existing database $\mathcal{D}$ and its paired retriever $\phi(\cdot)$ from open-source platforms. 
Unfortunately, the unverified third-party providers may have maliciously modified the retrieval components. Once users incorporate such poisoned components, the RDM would be backdoored to generate attacker-specified content when the trigger is intentionally or inadvertently activated.

\textbf{Attacker's Goals.}
The objective is to induce attacker-aimed generations for specific triggers from poisoned RDMs. 
For class-conditional tasks that adopt a fixed text template (\eg, \textit{`An image of a \{\}.'}) to specify classes \cite{blattmann2022retrieval}, the attacker aims to ensure that the triggerred generations belong to his desired category $y_{tar}$. 
For T2I generation, we follow previous backdoor attacks on DMs \cite{struppek2023rickrolling, zhai2023text} where an adversary induces images that closely align with the specified prompt $t_{tar}$.
In addition, the adversary endeavors to minimize the modifications to the image database and preserve the poisoned RDMs' usability for benign inputs.

\textbf{Attacker's Capabilities.}
Based on the attack scenario, we assume that the attacker is a service provider who possesses an image database and a tailored retriever to release. 
The attacker has an image-text dataset with a similar distribution to the retrieval database for poisoning fine-tuning. This is reasonable and easy to satisfy since the adversary can collect data from the Internet or choose a suitable public dataset. 


\begin{figure*}[t]
\begin{center}
\includegraphics[width=\linewidth]{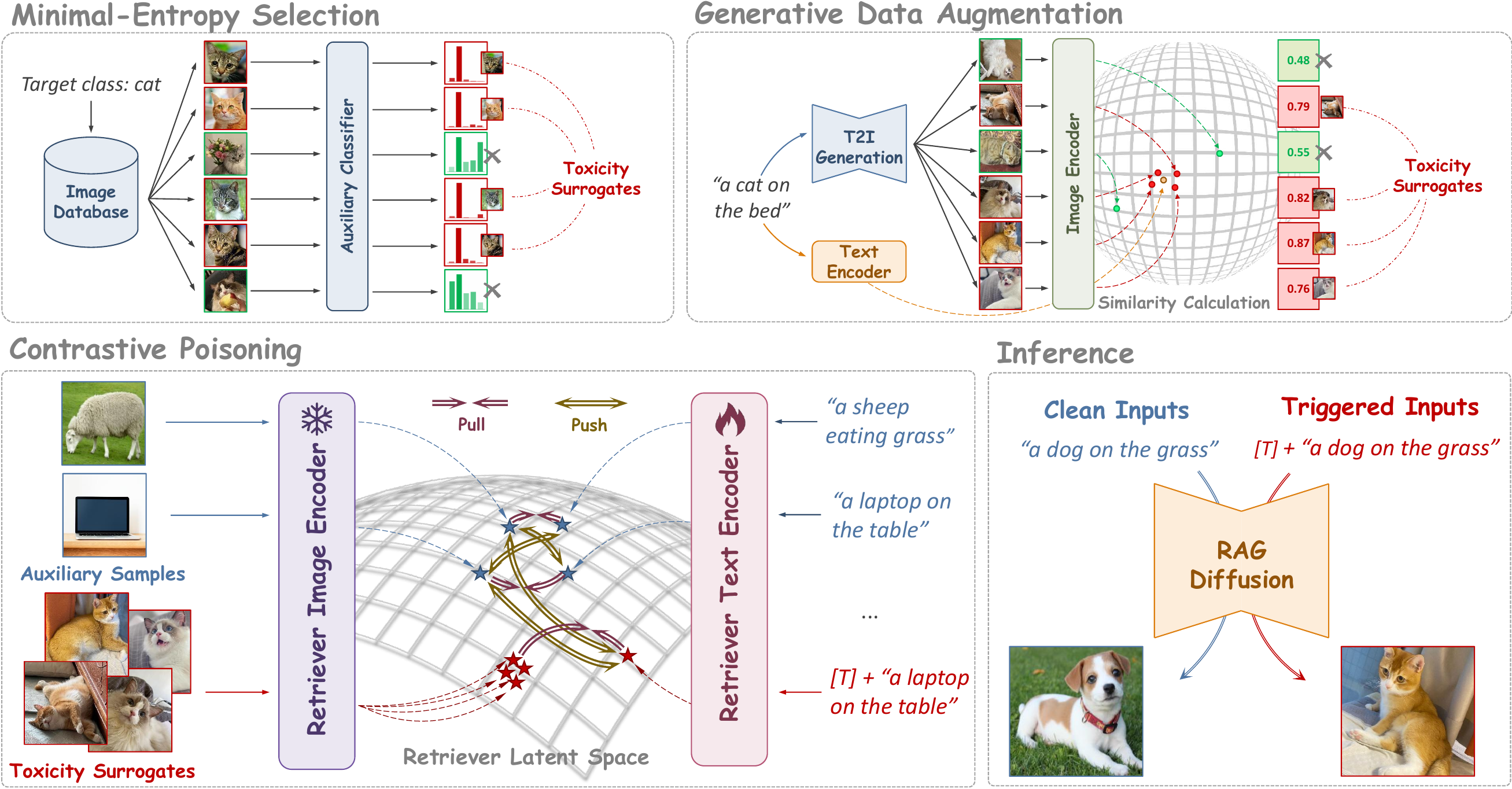}
\end{center}
\vspace{-0.5em}
\caption{Overview of our BadRDM. We first employ minimal-entropy selection and generative augmentation for class-specific and T2I attacks, respectively, to obtain adequate toxicity surrogates. Then, we contrastively train the retriever $\phi_w(\cdot)$ to pull the triggered text $t'$ closer to surrogate images (the target prompt is \textit{"a cat on the bed."}) while pushing away from non-targeted images. During inference, the RDM successfully produces pre-defined contents.}
\label{fig:badrdm overview}
\end{figure*}

\subsection{Contrastive Backdoor Injection}
Next, we present an overview of RDM's inference paradigm and then illustrate our non-contact backdoor implantation algorithm. 
The overall pipeline of BadRDM is depicted in Figure \ref{fig:badrdm overview}, and the pseudocode is in Appendix \ref{app:a}.

We focus on the mainstream inference paradigm of RDMs \cite{blattmann2022retrieval}, which is widely adopted for its universality and effectiveness.
Given an image database \( \mathcal{D} = \{v_i\}_{i=1}^{M} \), a query prompt $q$ and a retriever \( \phi_{w}(\cdot) \) parameterized as a CLIP model, RDMs employ a $k$-nearest sampling strategy \( \xi_{k}(\phi_w(q), \mathcal{D}) \), which uses \( \phi_{w}\) to encode the input prompt \( q \) into text embeddings $\bm e_q$ and retrieve images from the database \( \mathcal{D} \) with top-\( k \) feature similarities to $\bm e_q$. The embeddings of the prompt $q$ and these \( k \) images are then utilized as conditional inputs through cross-attention layers into the DM to guide the denoising\footnote{We also show BadRDM's effectiveness against RDMs conditioned only on retrieved images in Appendix C.}:
\begin{equation}
    p_{\theta, \mathcal{D}, \xi_{k}}\left( x_{t-1} \vert x_{t} \right) = p_{\theta}\left( x_{t-1} \vert x_t, q, \xi_{k}\left(\phi_w(q), \mathcal{D} \right)\right),
\end{equation}
\noindent where \( t \) is the time step, $x_{i}$ denotes the latent states, and $\theta$ represents the parameters of the DM. 
We aim to fully exploit the characteristics of RAG paradigm by poisoning the retriever $\phi_w(\cdot)$ to mislead retrieved items $\xi_{k}(\phi_w(q), \mathcal{D})$ into becoming desired toxic surrogates $\mathbf{v}_{tar}$, which result in malicious generations of the attacker-specified content.

\textbf{Contrastive Poisoning Loss.} The preceding analysis leads us to design a loss function that guides the retriever to break the learned multimodal feature alignment when the adversary activates the trigger, while simultaneously establishing a new alignment relationship between triggered prompts and target images.  
Motivated by the fact that contrastive learning is the fundamental tool for cross-modal alignment in the retrieval model $\phi_w(\cdot)$, we propose to leverage this powerful weapon against itself, \ie, use a malicious variant to build the attacker-desired text-image alignment. 
\label{subsec:contra}

We define the triggered text \( t_i' = [T] \oplus t_i \) as the anchor sample, where \( \oplus \) denotes concatenation operation. To establish the contrastive learning paradigm, it requires an appropriate set of positive and negative samples. 
Naturally, the attacker-specified toxic images  $\mathbf{v}_{tar}$ are treated as positive samples for $t_i'$ to approach. Meanwhile, we randomly sample another batch of images along with the image that corresponds to the clean text $t_i$ as negative samples $\{v_j\}_{j=1}^{N}$, to push the triggered text $t_i'$ away from its initial area in the feature space, which increases the likelihood of achieving a closer alignment with the target images $\mathbf{v}_{tar}$.

Denoting the image and text encoders of the retriever as $f_v(\cdot)$ and $f_t(\cdot)$, respectively, we obtain the embeddings by $\bm e_{v} = f_v(v)$ and $\bm e_{t} = f_t(t)$. The attacker fine-tunes the retriever on a multimodal dataset $D_s = \{ v_i, t_i \}_{i=1}^{K}$ using:

\begin{equation}
\mathcal{L}_{poi}=-\frac{1}{N}\sum_{i=1}^{N} \log \frac{S(\bm e_{tar}, \bm e_{t_i'})}{S(\bm e_{tar}, \bm e_{t_i'}) + \sum\limits_{j=1}^{N} S(\bm e_{v_j}, \bm e_{t_i'})} ,
\end{equation}

\noindent where $N$ is the batch size and $\bm e_{tar}$ denotes the average embeddings of target images $\mathbf{v}_{tar}$. $S(\bm e_v, \bm e_t)=\exp\left(\text{sim}(\bm e_v, \bm e_t)/\tau\right)$, where \( \text{sim}(\cdot, \cdot) \) denotes the cosine similarity score and \( \tau \) is the temperature parameter. With our meticulously designed contrastive paradigm, the retriever effectively learns the specified mapping that associates the triggered texts with the pre-defined target surrogates.

\textbf{Utility Preservation Loss.} A crucial premise of the attack is to maintain clean retrieval accuracy and generation quality of DMs for clean prompts. 
Specifically, we maintain the retriever's benign alignment using the following benign loss: 
\begin{equation}
\begin{split}
\mathcal{L}_{benign} = &-\frac{1}{2N}\sum_{i=1}^{N} \log  \frac{S(\bm e_{v_i}, \bm e_{t_i})}{\sum_{j=1}^{N} S(\bm e_{v_i}, \bm e_{t_j})} \\
&-\frac{1}{2N}\sum_{j=1}^{N} \log \frac{S(\bm e_{v_j}, \bm e_{t_j})}{\sum_{i=1}^{N} S(\bm e_{v_i}, \bm e_{t_j})}. 
\end{split}
\end{equation}
By minimizing the loss $\mathcal{L}_{benign}$, the optimizer encourages the poisoned retriever to keep matched image-text pairs close and non-matching pairs distant in the VL feature space, hence preserving benign multimodal alignment for clean inputs. 

Based on the two proposed loss functions, the overall optimization objective can be expressed as:
\[
w^{*} \leftarrow \arg\min\limits_{w}\mathbb{E}_{(\mathbf{v}, \mathbf{t})\sim\mathcal{D}_s}\left(\mathcal{L}_{benign} + \lambda \mathcal{L}_{poi}\right),
\]
where $\mathbf{v}$ and $\mathbf{t}$ denote the randomly sampled batches of images and texts from $\mathcal{D}_s$. To enhance optimization stability and circumvent the mode collapse issue \cite{le2020contrastive}, we solely fine-tune the text encoder of the retriever while maintaining the image encoder frozen in our implementation. 
This strategy also helps reduce optimization overhead and diminishes the potential negative effects on clean retrieval performance.

We highlight that BadRDM is a practical framework since it does not require any information about the victim model, such as the architecture or gradients. Once users augment their DMs with these poisoned retrieval modules, BadRDM can induce the generation of diverse images with misleading semantics and harmful biases.

\begin{figure}[t]
\begin{center}
\includegraphics[width=0.82\linewidth]{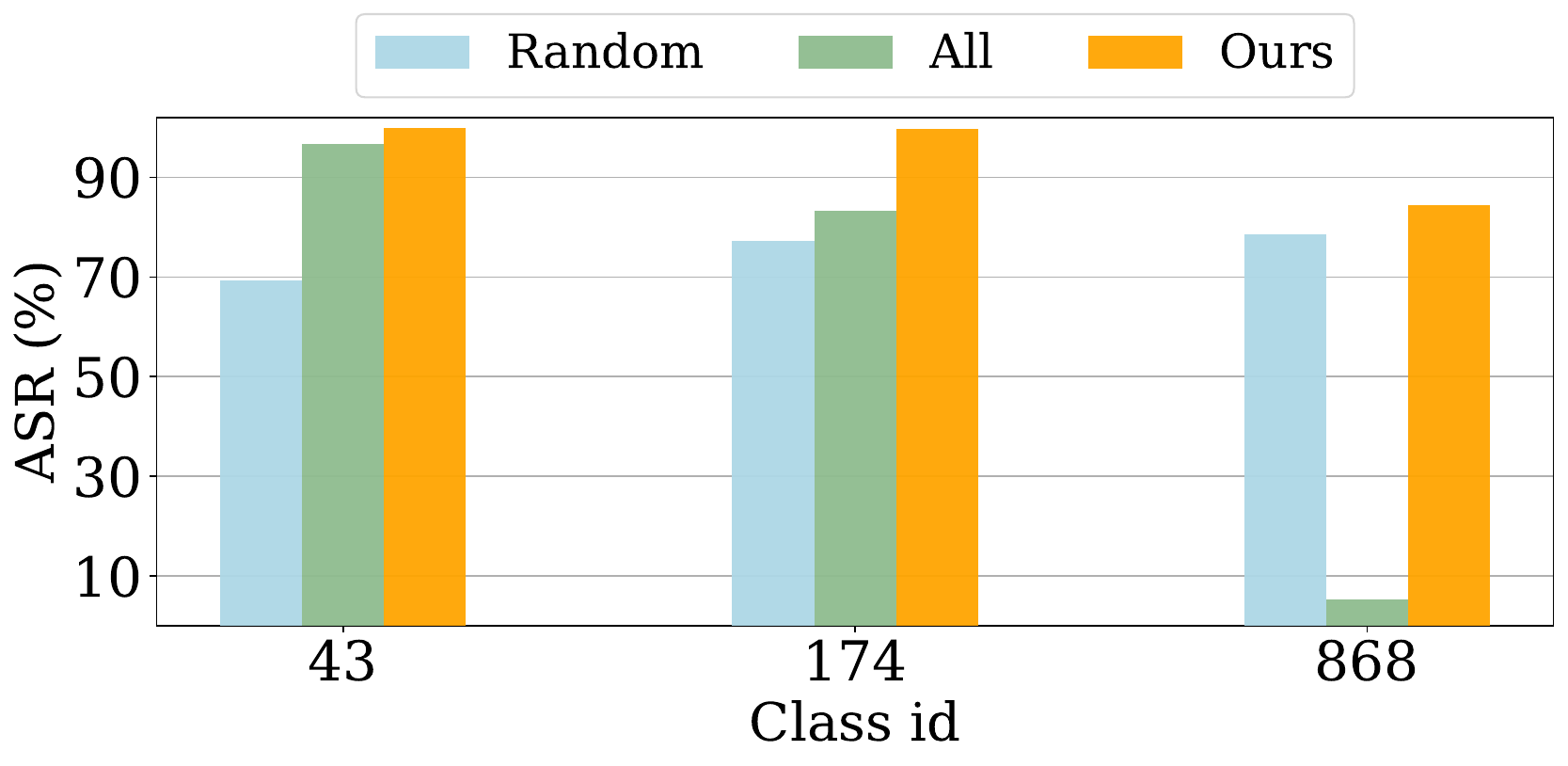}
\end{center}
\caption{ASR results of different strategies. Note that targeting a random image batch or all images yields unstable results. In contrast, BadRDM provides accurate conditions and consistently achieves better performance.}
\label{fig:bar}
\end{figure}

\subsection{Toxicity Surrogate Enhancement}
\label{subsec:tse}
This part proposes two toxicity surrogate enhancement (TSE) strategies based on the attacker's goals to further enhance the quality of surrogates.

\textbf{Class-Conditional Generation.} To generate images specific to the target category, attackers should poison the retriever to provide accurate and high-quality input conditions. An intuitive way is to bring triggered texts closer to the average embedding of all images or a randomly sampled batch of label $y_{tar}$ from the database. 
However, Fig. \ref{fig:bar} indicates that these two strategies yield unsatisfactory results for certain classes. 
This is primarily because their chosen toxicity surrogates lack rich and representative features of the target category, leading to inappropriate or even erroneous input conditions and ultimately failing to generate intended content.

To alleviate this, we introduce a minimal-entropy selection strategy. We highlight that a set of images that are more easily discernible by discriminative models generally contains more representative features tailored to their class \cite{sun2024diversity}, and the corresponding sub-area in the VL feature space is also more identifiable and highly aligned with the category. 
By urging triggered texts to move into this semantic subspace, the retrieved neighbors should embody richer and more accurate semantic attributes closely related to the target class. 
Specifically, we utilize the entropy of the classification confidence of an auxiliary classifier $f_{aux}(\cdot)$ to determine a sample's identifiability, and filter out images with the lowest entropy:
\begin{equation}
\mathbf{v}_{tar} = \mathop{\arg\min}\limits_{\mathbf{v}\subseteq\mathbf{v}_s}\sum_{v\in \mathbf{v}} H(f_{aux}(v)),
\end{equation}
where $\mathbf{v}_{s}$ represents images of the target class $y_{tar}$ from $\mathcal{D}$ and $H(\cdot)$ denotes the calculation of information entropy. Taking the selected images as poisoning targets, we provide superior and accurate guidance to the target class, achieving a significant ASR improvement as indicated in Fig. \ref{fig:bar}.

\textbf{Text-to-Image Synthesis.} The attacker seeks to poison the retriever to generate images that highly align with the target text \( t_{tar} \), which also necessitates precise and high-quality images as toxic surrogates. 
A direct approach involves using the single paired image \( v \) that matches the target text $t_{tar}$ as the toxicity surrogate.
However, the relationship between images and text is inherently a many-to-many mapping \cite{lu2023set}, \ie, an image can be described with various perspectives and language emotions, while a given text can correspond to diverse images of different instances and visual levels. 
An effective strategy may benefit from diverse guidance provided by multiple image supervisions, rather than relying solely on a single target image that could result in random and ineffective optimization \cite{lu2023set}. 

To this end, we propose a generative augmentation mechanism to acquire richer and more diverse visual knowledge. Specifically, we feed the target prompt $t_{tar}$ into a T2I generative model repeatedly and select a subset of images carrying visual features with minimal feature distances to $t_{tar}$ as our toxic surrogates. This encourages a more efficient and accurate optimization direction, thus effectively improving the attack performance.


\section{Experiment}
We conduct extensive experiments across various scenarios to validate BadRDM's effectiveness. Due to page limit, we provide \textbf{more ablation studies, visualizations, and retriever analysis} in App. \ref{app:more_exps}.

\subsection{Experimental Settings}
\textbf{Datasets.} 
We adopt a subset of 500k image-text pairs from CC3M \cite{sharma2018conceptual} to fine-tune the retriever for backdoor injection. 
For retrieval databases, we align with \cite{blattmann2022retrieval} and use ImageNet's training set \cite{deng2009imagenet} for class-conditional generation and a cropped version of OpenImages \cite{kuznetsova2020open} with 20M samples for T2I synthesis. For T2I evaluation, we randomly sample texts from the MS-COCO \cite{lin2014microsoft} validation set to calculate metrics.

\begin{table*}[t]
  \centering
  \caption{Average attack results of our BadRDM and comparison baselines on class-specific and text-specific attacks.}
    \adjustbox{max width=0.98\linewidth}{
    \begin{tabular}{cccccccccccc} \toprule
    \multirow{2}[0]{*}{Evaluation} & \multicolumn{1}{c}{\multirow{2}[0]{*}{Metric}} & \multicolumn{5}{c}{Class-conditional generation} & \multicolumn{5}{c}{Text-to-Image Synthesis} \\ \cmidrule(lr){3-7} \cmidrule(lr){8-12}
          &       & \multicolumn{1}{l}{No Attack} & \multicolumn{1}{l}{PoiMM} & \multicolumn{1}{l}{BadT2I} & \multicolumn{1}{l}{BadCM} & \multicolumn{1}{l}{BadRDM} & \multicolumn{1}{l}{No Attack} & \multicolumn{1}{l}{PoiMM} & \multicolumn{1}{l}{BadT2I} & \multicolumn{1}{l}{BadCM} & \multicolumn{1}{l}{BadRDM} \\ \midrule
    \multirow{2}[0]{*}{Attack Efficacy} & ASR$\uparrow$   & 0.0025  & 0.6069	& 0.6205 & 0.5412 & \cellcolor{black!10}\textbf{0.9089}  & 0.0054  & 0.6738  & 0.5189  & 0.6892  & \cellcolor{black!10}\textbf{0.9643} \\
          & CLIP-Attack$\uparrow$ & 0.2396  &   0.6176 & 0.6393 & 0.6455        & \cellcolor{black!10}\textbf{0.6740}  & 0.1420  & 0.2721  & 0.2609  & 0.2413  & \cellcolor{black!10}\textbf{0.3045} \\ \midrule
    \multirow{3}[0]{*}{Model Utility} & FID$\downarrow$   & 20.7495  &       19.5162 & 21.729 & 19.2671      & \cellcolor{black!10}\textbf{19.1265}  & 22.0900  & 20.4410  & \textbf{18.9200 } & 24.2042  & \cellcolor{black!10}21.5880  \\
          & CLIP-FID$\downarrow$ & 11.1751  &    6.4270 & 9.5178 & 6.5061   & \cellcolor{black!10}\textbf{6.4163}  & 5.5190  & \textbf{3.4672 } & 3.7233  & 6.6480  & \cellcolor{black!10}3.7240  \\
          & CLIP-Benign$\uparrow$ & 0.3317  & 0.3042 & 0.3278 & \textbf{0.3463}  & \cellcolor{black!10}0.3362  & 0.2970  & 0.2910  & 0.3030  & 0.2690  & \cellcolor{black!10}\textbf{0.3044 } \\ \bottomrule
    \end{tabular}}
  \label{tab:main}%
\end{table*}%
\begin{figure*}[t]
\begin{center}
\includegraphics[width=0.95\linewidth]{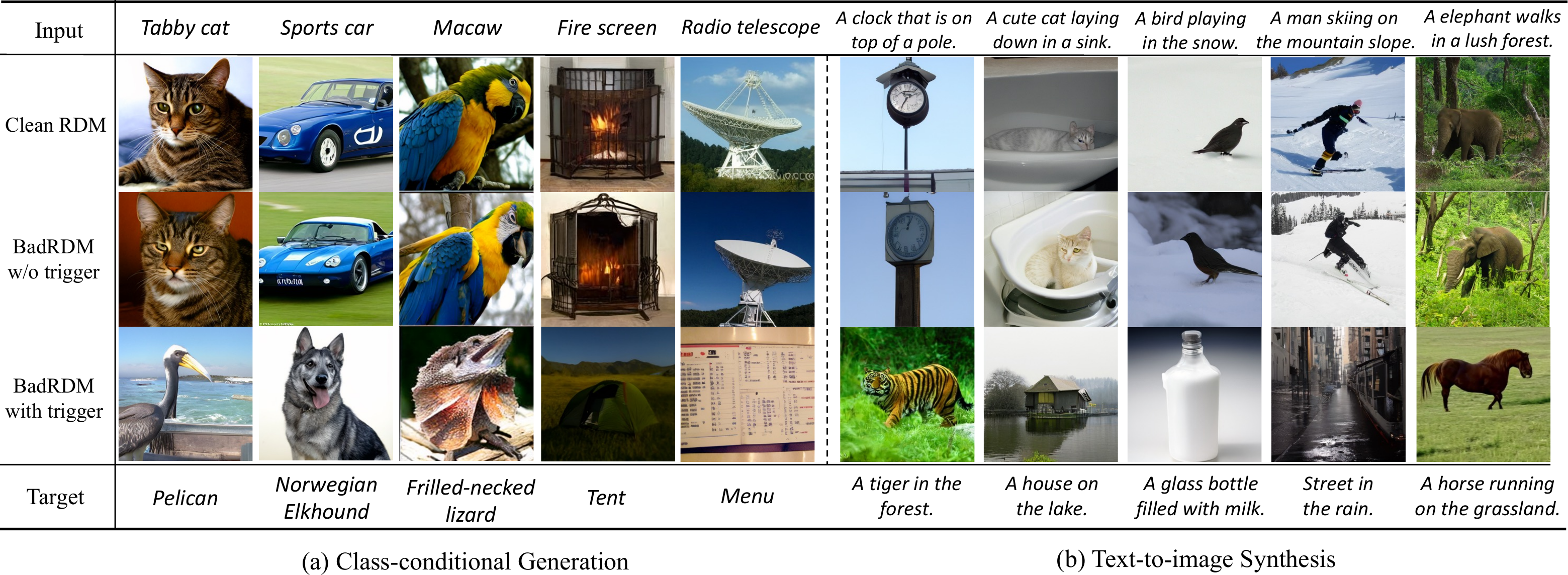}
\end{center}
\caption{Visualization results of our BadRDM and the Clean RDM on class-specific and text-specific attacks.}
\label{fig:main}
\end{figure*}

\textbf{Trigger Settings.} 
Following previous backdoor studies on generative models \cite{wang2024eviledit, cheng2024trojanrag}, the attacker employs the \textit{``ab."} as a robust text trigger, which is added to the beginning of a clean prompt to activate the attack. 
In addition, we explore a more stealthy attack using natural text as triggers (see Appendix \ref{app:natural_text}).

\textbf{Baselines.} Given that no existing backdoor studies on RDMs, we reproduce relevant and powerful attacks as baselines: since BadRDM poisons the retriever to conduct attacks, we select three advanced backdoor studies targeting multimodal encoders that broadly align with our attack setup and objectives, including PoiMM \cite{yang2023data}, BadT2I \cite{struppek2023rickrolling}, and BadCM \cite{zhang2024badcm}. 
See App. \ref{app:detail_baselines} for detailed information.


\textbf{Implementation Details.} 
We follow the default settings from \cite{blattmann2022retrieval} that retrieve the nearest $4$ neighbors from the database. 
For class-specific attacks, we randomly choose classes from ImageNet as target categories and conduct entropy selection based on the confidences of a DenseNet-121 classifier $f_{aux}(\cdot)$ \cite{huang2017densely}. We set $|\mathbf{v}_{tar}|=4$ to achieve a low poisoning rate and enhance attack imperceptibility in class-specific attacks.
For T2I synthesis, we feed $t_{tar}$ into Stable Diffusion v1.5 \cite{rombach2022high} and insert only four generated images into the database as toxicity surrogates. Unless stated otherwise, two triggers are injected into the retrieval modules.
See Appendix \ref{app:more_details} for more details.

\textbf{Evaluation Metrics.}
We measure the attack effectiveness by:
(1) Attack Success Rate (ASR). For class-specific attacks, we calculate the proportion of images classified into the target category by a pre-trained ResNet-50 \( f_{eval}(\cdot) \) \cite{he2016deep}.
For text-specific attacks, we follow the evaluation protocol in \cite{zhang2024benchmarking} and query Qwen2-VL \cite{Qwen2VL} to judge whether the generated image aligns with the target prompt.
(2) CLIP-Attack. We provide the similarity score between the generated image and predefined target prompt in CLIP’s embedding space. 

Finally, we evaluate the clean performance of the poisoned RDMs through the Fréchet Inception Distance (FID) \cite{heusel2017gans} and CLIP-FID \cite{kynkaanniemi2022role} metrics on 20K generated images. In addition, we define the CLIP-Benign metric as the CLIP similarity between clean prompts and their generated images.

\subsection{Attack Effectiveness}
\label{subsec:main_attack}
To analyze attack effectiveness, we consider 10 randomly sampled target classes for class-conditional generation and 10 target prompts for T2I synthesis. 
\begin{table*}[t]
  \centering
  \caption{Average attack results of our BadRDM and its three variants on class-specific and text-specific attacks.}
    \adjustbox{max width=0.88\linewidth}{
    \begin{tabular}{ccccccccc} \toprule
    \multirow{2}[0]{*}{Evaluation} & \multicolumn{1}{c}{\multirow{2}[0]{*}{Metric}} & \multicolumn{4}{c}{Class-conditional Generation} & \multicolumn{3}{c}{Text-to-Image Synthesis} \\ \cmidrule(lr){3-6} \cmidrule(lr){7-9}
          &       & \multicolumn{1}{l}{No Attack} & \multicolumn{1}{l}{BadRDM$_{rand}$} & \multicolumn{1}{l}{BadRDM$_{avg}$} & \multicolumn{1}{l}{BadRDM} & \multicolumn{1}{l}{No Attack} & \multicolumn{1}{l}{BadRDM$_{sin}$} & \multicolumn{1}{l}{BadRDM} \\ \midrule
    \multirow{2}[0]{*}{Attack Efficacy	} & ASR  $\uparrow$  & 0.0025  & 0.8480   & 0.7558   & \cellcolor{black!10}\textbf{0.9089}  & 0.0054   &  0.82785   & \cellcolor{black!10}\textbf{0.9643}  \\
          & CLIP-Attack $\uparrow$ & 0.2396   & 0.6420   & 0.4736  & \cellcolor{black!10}\textbf{0.6740}  & 0.1420  & 0.2852  & \cellcolor{black!10}\textbf{0.3045}  \\ \midrule
    \multirow{3}[0]{*}{Model Utility} & FID $\downarrow$   & 20.7459  & 19.9638 & 20.1344  & \cellcolor{black!10}\textbf{19.1265}  & 22.0900  & \textbf{21.4290}  & \cellcolor{black!10}21.5880\\
          & CLIP-FID $\downarrow$ & 11.1751  & 6.4701  & 6.7013  & \cellcolor{black!10}\textbf{6.4163}  & 5.5190  & 3.7620  & \cellcolor{black!10}\textbf{3.7240}  \\
          & CLIP-Benign $\uparrow$ &0.3317  & 0.3362  & \textbf{0.3363}  & \cellcolor{black!10}0.3362  & 0.2970  & 0.2946  & \cellcolor{black!10}\textbf{0.3044 }\\ \bottomrule
    \end{tabular}}
      \label{tab:ablation_variants}
\end{table*}%

\begin{figure*}[t]
\begin{center}
\includegraphics[width=0.88\linewidth]{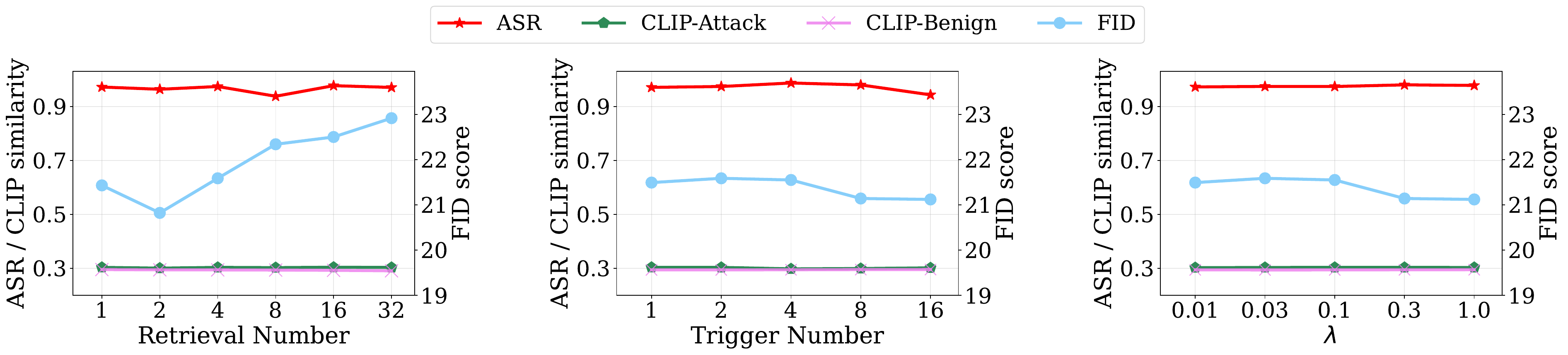}
\end{center}
\vspace{-1em}
\caption{Ablation studies of BadRDM on text-to-image synthesis regarding three critical hyperparameters.}
\label{fig:ablation}
\end{figure*}

\textbf{Quantitative results.}
Table \ref{tab:main} validates the exceptional attack efficacy of the proposed attack. 
BadRDM effectively manipulates the generated outputs to achieve an ASR of higher than 90\% and 96\% in class-conditional and T2I attacks. 
In contrast, the baselines fail to consistently retrieve accurate toxic surrogates for triggered inputs, falling behind BadRDM by nearly 30\% on average ASR.
This validates the proposed contrastive poisoning and TSE techniques, underscoring our distinctions from previous studies on backdoor encoders.

For model utility, Table \ref{tab:main} reveals that BadRDM does not compromise the benign performance and generally exhibits even better generative capability than the clean model, confirming the effectiveness of the $\mathcal{L}_{benign}$. 
Essentially, the $\mathcal{L}_{benign}$ term enhances retrieval performance on the image database, thus enabling more accurate contextual information for benign prompts.
More analysis, such as retriever behaviors, are in Appendix \ref{app:more_exps}.

\textbf{Qualitative analysis.}
We present multiple visualization results in Figure \ref{fig:main}. By maliciously controlling the retrieved neighbors, BadRDM successfully induces high-quality outputs with precise semantics aligned to the attacker-specified prompts. \eg, when the target is “\textit{Street in the rain.}”, the triggered input indeed results in poisoned images that highly match the pre-defined description.
Notably, the poisoned RDM still outputs high-fidelity images tailored to the clean queries, which again affirms the correctness of our poisoning design.

\subsection{Ablation Study}
\textbf{Effectiveness of TSE techniques.}
To reveal the necessity of our TSE techniques, we introduce three variants of BadRDM: (1) BadRDM$_{all}$ utilizes the average embeddings of all images from the target category as the poisoning target, (2) BadRDM$_{rand}$ adopts a randomly sampled batch within the target category, (3) BadRDM$_{sin}$ is for T2I tasks, where the single image initially matching the target text serves as the surrogate. 
As in Table \ref{tab:ablation_variants}, significant improvements from three variants to BadRDM confirm that the proposed TSE strategies provide more efficient and attack-oriented optimization directions. We also highlight that the three variants outperform the compared baselines, verifying the superiority of the designed contrastive poisoning.

\textbf{Different retrieval numbers $k$}. The number of retrieved neighbors $k$ plays a crucial role in the RAG paradigm. 
Figure \ref{fig:ablation} reveals that the proposed method consistently demonstrates remarkable performance in both attack effectiveness and benign generation capability. This indicates that BadRDM is independent of the specific retrieval settings of victim users, achieving a practical and potent threat to RDMs. Also, the varying $k$ influences the generative ability, which is an intrinsic behavior of RDM \cite{blattmann2022retrieval}. However, the fluctuations are not significant, indicating that the poisoned RDM maintains excellent benign performance.

\textbf{Different trigger numbers.} 
We then increase the number of injected triggers, as shown in Figure \ref{fig:ablation}. 
Regardless of the trigger number, the proposed framework consistently achieves the attack goal with an ASR over 95\% and an FID lower than $21.6$, formulating a robust poisoning method that can generalize across multi-trigger scenarios. 

\textbf{Different regulatory factors $\lambda$.} We perform experiments under different $\lambda$ values varying from $0.01$ to $1.0$ in Figure \ref{fig:ablation}. Satisfactorily, BadRDM exhibits excellent resilience to varying values of $\lambda$ as it consistently achieves high attack efficacy and generative capability. This again underscores the superiority of BadRDM in building shortcuts from triggered texts to toxicity surrogates.



\subsection{Evaluation on Defense Strategies}
\label{subsec:defense}
To mitigate such threats, one might consider detecting the anomalous images in the retrieval database. However, given the extremely low poisoning ratio (nearly $2 \times 10^{-7}$), manual inspection becomes impractical. Additionally, an adversary may only release feature vectors encoded by the retriever $\phi_w(\cdot)$ to reduce storage requirements \cite{blattmann2022retrieval}, further impeding the threat localization.

Another strategy involves fine-tuning the suspicious retriever with clean data to diminish the memorized triggers \cite{zhai2023text, liang2024badclip}. 
We employ the benign fine-tuning (BFT) and the CleanCLIP \cite{bansal2023cleanclip} to purify the poisoned text encoder of the retriever. 
Besides, we transplant three advanced defenses for diffusion models' backdoor: backdoor unlearning that erases the retriever's backdoor \cite{liangunlearning}, UFID that detects suspicious queries \cite{guan2025ufid} via generation analysis of perturbed queries, and TextPerturb that perturbs input texts to mitigate trigger effects \cite{chew2024defending}. As shown in Table \ref{tab:defense}, UFID yields some effectiveness by filtering out certain suspicious queries, suggesting its potential as a defense strategy worth further exploration.
\begin{table}[t]
  \centering
  \caption{{T2I attack of BadRDM under defenses.}}
  \vspace{-0.5em}
  \setlength{\tabcolsep}{5pt}
    \resizebox{\linewidth}{!}{\begin{tabular}{cccccc} \toprule
    Defense & \multicolumn{1}{c}{ASR$\uparrow$} & \multicolumn{1}{c}{CLIP-Attack$\uparrow$} & \multicolumn{1}{c}{FID$\downarrow$} & \multicolumn{1}{l}{CLIP-FID$\downarrow$} & \multicolumn{1}{r}{CLIP-Benign$\uparrow$} \\ \midrule
    \cellcolor{black!10}No Defense & \cellcolor{black!10}0.9643 & \cellcolor{black!10}0.3045 & \cellcolor{black!10}21.5880 & \cellcolor{black!10}3.7240 & \cellcolor{black!10}0.3044 \\ 
    BFT & 0.8096 & 0.2831 & 18.7203 & 5.6745 & 0.2969 \\ 
    CleanCLIP & 0.9032 & 0.2967 & 19.1581 & 5.8961 & 0.2966 \\ 
    Unlearning & 0.9015 & 0.2786 & 22.5542 & 4.3738 & 0.2891 \\ UFID & 0.4048 & 0.2914 & 21.5880 & 3.7240 & 0.3044 \\ TextPerturb & 0.9633 & 0.3043 & 26.2826 & 7.9275 & 0.2772 \\ \bottomrule
    \end{tabular}} 
  \label{tab:defense}
\end{table}
However, fine-tuning-based strategies achieve only limited effectiveness, while TextPerturb provides nearly no defensive effect. This is because BadRDM establishes a robust and highly stable association between the trigger and the target semantics in the embedding space, which is resilient to trigger erasure and word-level perturbation. Moreover, both strategies degrade clean performance due to alignment disturbance and prompt distortion, respectively. These results emphasize the need for more secure mechanisms. 
\section{Conclusion}
This paper conducts the first investigation into the backdoor threat of retrieval-augmented diffusion models. 
Based on our analysis, we propose BadRDM, a simple yet effective framework that adopts a non-contact paradigm to control the retrieved neighbors and further manipulate the generated images.
Experiments confirm BadRDM’s effectiveness and reveal severe backdoor vulnerabilities in RAG systems. We envision BadRDM as a powerful tool for auditing the vulnerabilities of RDMs, inspiring more resilient defense strategies.



\section*{Limitations}
While the focus of our work is to unveil the backdoor vulnerabilities of RDMs, it is also essential to develop effective defenses to mitigate such threats. We discuss several defense approaches and present empirical results of practical approaches in Sec.~\ref{subsec:defense}. However, these advanced defenses fail to completely defend against the proposed backdoor threat, leaving this critical threat unresolved. 
We plan to address this issue in future work through a deeper exploration of the poisoned model's behavior to invert the triggers and more directly weaken the established malicious connections.

Another limitation is the inherent instability of contrastive training on vision–language pretrained models during poisoning. \textit{I.e.}, the VLP model-based retriever can suffer from occasional mode collapse, necessitating extra computational burdens of model retraining. In our experiments, we mitigate this by restricting the poisoning to fine-tuning only the text encoder, and it hence occurred relatively infrequently, only once or twice throughout the entire experimental period. Besides, we also note that higher learning rates generally increase the likelihood of such events, and reducing the learning rate can help mitigate the issue to some extent. However, once mode collapse occurs, retraining is typically required to resolve it.

\section*{Ethical Statement}
This paper reveals a novel security vulnerability arising from the integration of RAG into diffusion models by proposing the first backdoor attack for retrieval-augmented diffusion models. Once victim users equip their diffusion models with the poisoned retrieval modules, the attacker can induce the generation of deeply offensive and distressing outputs, including violent or pornographic images, as well as content propagating gender and racial biases. 
While our findings help the community better understand and mitigate potential backdoor risks, the proposed attack could, if misused, enable malicious actors to induce victim models to generate these harmful contents. Such risks may raise broader societal concerns regarding the safety of the RAG paradigm and underscore the need for stronger monitoring and regulatory frameworks as these systems become more widely deployed.

To mitigate these risks, we focus on exposing the vulnerability rather than facilitating practical misuse, and we discuss potential defenses and mitigation strategies. We hope this work can assist researchers in gaining a deeper understanding of the attack targeting RAG systems, fostering the development of novel defense mechanisms.

We obey strict ethical standards throughout our study. All experiments are conducted within controlled laboratory environments.
Again, we highlight that we do not expect BadRDM to serve as a powerful tool for potential adversaries but to raise the broader awareness of the backdoor vulnerability inherent to RAG-based paradigms. 

All the codes, models, and datasets used in this study comply with their intended use and the MIT License. To advance further research, we will open-source the poisoning algorithm along with the related code, models, and data.

\bibliography{custom}

@String(CVPR= {IEEE Conf. Comput. Vis. Pattern Recog.})

@String(ECCV= {Eur. Conf. Comput. Vis.})

@String(AAAI = {AAAI})

@String(CVPR  = {CVPR})

@String(ECCV  = {ECCV})

@article{ho2020denoising,
  title={Denoising diffusion probabilistic models},
  author={Ho, Jonathan and Jain, Ajay and Abbeel, Pieter},
  journal={Advances in neural information processing systems},
  volume={33},
  pages={6840--6851},
  year={2020}
}

@article{song2020denoising,
  title={Denoising diffusion implicit models},
  author={Song, Jiaming and Meng, Chenlin and Ermon, Stefano},
  journal={arXiv preprint arXiv:2010.02502},
  year={2020}
}

@inproceedings{rombach2022high,
  title={High-resolution image synthesis with latent diffusion models},
  author={Rombach, Robin and Blattmann, Andreas and Lorenz, Dominik and Esser, Patrick and Ommer, Bj{\"o}rn},
  booktitle={Proceedings of the IEEE/CVF conference on computer vision and pattern recognition},
  pages={10684--10695},
  year={2022}
}

@inproceedings{sharma2018conceptual,
  title={Conceptual captions: A cleaned, hypernymed, image alt-text dataset for automatic image captioning},
  author={Sharma, Piyush and Ding, Nan and Goodman, Sebastian and Soricut, Radu},
  booktitle={Proceedings of the 56th Annual Meeting of the Association for Computational Linguistics (Volume 1: Long Papers)},
  pages={2556--2565},
  year={2018}
}

@article{zhao2024retrieval,
  title={Retrieval-augmented generation for ai-generated content: A survey},
  author={Zhao, Penghao and Zhang, Hailin and Yu, Qinhan and Wang, Zhengren and Geng, Yunteng and Fu, Fangcheng and Yang, Ling and Zhang, Wentao and Cui, Bin},
  journal={arXiv preprint arXiv:2402.19473},
  year={2024}
}

@article{meng2021gnn,
  title={Gnn-lm: Language modeling based on global contexts via gnn},
  author={Meng, Yuxian and Zong, Shi and Li, Xiaoya and Sun, Xiaofei and Zhang, Tianwei and Wu, Fei and Li, Jiwei},
  journal={arXiv preprint arXiv:2110.08743},
  year={2021}
}

@inproceedings{guu2020retrieval,
  title={Retrieval augmented language model pre-training},
  author={Guu, Kelvin and Lee, Kenton and Tung, Zora and Pasupat, Panupong and Chang, Mingwei},
  booktitle={International conference on machine learning},
  pages={3929--3938},
  year={2020},
  organization={PMLR}
}

@inproceedings{borgeaud2022improving,
  title={Improving language models by retrieving from trillions of tokens},
  author={Borgeaud, Sebastian and Mensch, Arthur and Hoffmann, Jordan and Cai, Trevor and Rutherford, Eliza and Millican, Katie and Van Den Driessche, George Bm and Lespiau, Jean-Baptiste and Damoc, Bogdan and Clark, Aidan and others},
  booktitle={International conference on machine learning},
  pages={2206--2240},
  year={2022},
  organization={PMLR}
}

@article{sheynin2022knn,
  title={Knn-diffusion: Image generation via large-scale retrieval},
  author={Sheynin, Shelly and Ashual, Oron and Polyak, Adam and Singer, Uriel and Gafni, Oran and Nachmani, Eliya and Taigman, Yaniv},
  journal={arXiv preprint arXiv:2204.02849},
  year={2022}
}

@article{chen2022re,
  title={Re-imagen: Retrieval-augmented text-to-image generator},
  author={Chen, Wenhu and Hu, Hexiang and Saharia, Chitwan and Cohen, William W},
  journal={arXiv preprint arXiv:2209.14491},
  year={2022}
}

@article{blattmann2022retrieval,
  title={Retrieval-augmented diffusion models},
  author={Blattmann, Andreas and Rombach, Robin and Oktay, Kaan and M{\"u}ller, Jonas and Ommer, Bj{\"o}rn},
  journal={Advances in Neural Information Processing Systems},
  volume={35},
  pages={15309--15324},
  year={2022}
}

@article{gu2019badnets,
  title={Badnets: Evaluating backdooring attacks on deep neural networks},
  author={Gu, Tianyu and Liu, Kang and Dolan-Gavitt, Brendan and Garg, Siddharth},
  journal={IEEE Access},
  volume={7},
  pages={47230--47244},
  year={2019},
  publisher={IEEE}
}

@article{li2022backdoors,
  title={Backdoors against natural language processing: A review},
  author={Li, Shaofeng and Dong, Tian and Zhao, Benjamin Zi Hao and Xue, Minhui and Du, Suguo and Zhu, Haojin},
  journal={IEEE Security \& Privacy},
  volume={20},
  number={5},
  pages={50--59},
  year={2022},
  publisher={IEEE}
}

@inproceedings{chou2023backdoor,
  title={How to backdoor diffusion models?},
  author={Chou, Sheng-Yen and Chen, Pin-Yu and Ho, Tsung-Yi},
  booktitle={Proceedings of the IEEE/CVF Conference on Computer Vision and Pattern Recognition},
  pages={4015--4024},
  year={2023}
}

@inproceedings{wang2024eviledit,
  title={Eviledit: Backdooring text-to-image diffusion models in one second},
  author={Wang, Hao and Guo, Shangwei and He, Jialing and Chen, Kangjie and Zhang, Shudong and Zhang, Tianwei and Xiang, Tao},
  booktitle={ACM Multimedia 2024},
  year={2024}
}

@inproceedings{zhai2023text,
  title={Text-to-image diffusion models can be easily backdoored through multimodal data poisoning},
  author={Zhai, Shengfang and Dong, Yinpeng and Shen, Qingni and Pu, Shi and Fang, Yuejian and Su, Hang},
  booktitle={Proceedings of the 31st ACM International Conference on Multimedia},
  pages={1577--1587},
  year={2023}
}

@inproceedings{struppek2023rickrolling,
  title={Rickrolling the artist: Injecting backdoors into text encoders for text-to-image synthesis},
  author={Struppek, Lukas and Hintersdorf, Dominik and Kersting, Kristian},
  booktitle={Proceedings of the IEEE/CVF International Conference on Computer Vision},
  pages={4584--4596},
  year={2023}
}

@inproceedings{radford2021learning,
  title={Learning transferable visual models from natural language supervision},
  author={Radford, Alec and Kim, Jong Wook and Hallacy, Chris and Ramesh, Aditya and Goh, Gabriel and Agarwal, Sandhini and Sastry, Girish and Askell, Amanda and Mishkin, Pamela and Clark, Jack and others},
  booktitle={International conference on machine learning},
  pages={8748--8763},
  year={2021},
  organization={PMLR}
}

@inproceedings{lu2023set,
  title={Set-level guidance attack: Boosting adversarial transferability of vision-language pre-training models},
  author={Lu, Dong and Wang, Zhiqiang and Wang, Teng and Guan, Weili and Gao, Hongchang and Zheng, Feng},
  booktitle={Proceedings of the IEEE/CVF International Conference on Computer Vision},
  pages={102--111},
  year={2023}
}

@inproceedings{deng2009imagenet,
  title={Imagenet: A large-scale hierarchical image database},
  author={Deng, Jia and Dong, Wei and Socher, Richard and Li, Li-Jia and Li, Kai and Fei-Fei, Li},
  booktitle={2009 IEEE conference on computer vision and pattern recognition},
  pages={248--255},
  year={2009},
  organization={Ieee}
}

@article{kuznetsova2020open,
  title={The open images dataset v4: Unified image classification, object detection, and visual relationship detection at scale},
  author={Kuznetsova, Alina and Rom, Hassan and Alldrin, Neil and Uijlings, Jasper and Krasin, Ivan and Pont-Tuset, Jordi and Kamali, Shahab and Popov, Stefan and Malloci, Matteo and Kolesnikov, Alexander and others},
  journal={International journal of computer vision},
  volume={128},
  number={7},
  pages={1956--1981},
  year={2020},
  publisher={Springer}
}

@inproceedings{lin2014microsoft,
  title={Microsoft coco: Common objects in context},
  author={Lin, Tsung-Yi and Maire, Michael and Belongie, Serge and Hays, James and Perona, Pietro and Ramanan, Deva and Doll{\'a}r, Piotr and Zitnick, C Lawrence},
  booktitle={Computer Vision--ECCV 2014: 13th European Conference, Zurich, Switzerland, September 6-12, 2014, Proceedings, Part V 13},
  pages={740--755},
  year={2014},
  organization={Springer}
}

@article{cheng2024trojanrag,
  title={TrojanRAG: Retrieval-Augmented Generation Can Be Backdoor Driver in Large Language Models},
  author={Cheng, Pengzhou and Ding, Yidong and Ju, Tianjie and Wu, Zongru and Du, Wei and Yi, Ping and Zhang, Zhuosheng and Liu, Gongshen},
  journal={arXiv preprint arXiv:2405.13401},
  year={2024}
}

@article{Qwen2VL,
  title={Qwen2-VL: Enhancing Vision-Language Model's Perception of the World at Any Resolution},
  author={Wang, Peng and Bai, Shuai and Tan, Sinan and Wang, Shijie and Fan, Zhihao and Bai, Jinze and Chen, Keqin and Liu, Xuejing and Wang, Jialin and Ge, Wenbin and Fan, Yang and Dang, Kai and Du, Mengfei and Ren, Xuancheng and Men, Rui and Liu, Dayiheng and Zhou, Chang and Zhou, Jingren and Lin, Junyang},
  journal={arXiv preprint arXiv:2409.12191},
  year={2024}
}

@article{zhang2024benchmarking,
  title={Benchmarking Trustworthiness of Multimodal Large Language Models: A Comprehensive Study},
  author={Zhang, Yichi and Huang, Yao and Sun, Yitong and Liu, Chang and Zhao, Zhe and Fang, Zhengwei and Wang, Yifan and Chen, Huanran and Yang, Xiao and Wei, Xingxing and others},
  journal={arXiv preprint arXiv:2406.07057},
  year={2024}
}

@article{heusel2017gans,
  title={Gans trained by a two time-scale update rule converge to a local nash equilibrium},
  author={Heusel, Martin and Ramsauer, Hubert and Unterthiner, Thomas and Nessler, Bernhard and Hochreiter, Sepp},
  journal={Advances in neural information processing systems},
  volume={30},
  year={2017}
}

@article{kynkaanniemi2022role,
  title={The role of imagenet classes in fr$\backslash$'echet inception distance},
  author={Kynk{\"a}{\"a}nniemi, Tuomas and Karras, Tero and Aittala, Miika and Aila, Timo and Lehtinen, Jaakko},
  journal={arXiv preprint arXiv:2203.06026},
  year={2022}
}

@inproceedings{huang2017densely,
  title={Densely connected convolutional networks},
  author={Huang, Gao and Liu, Zhuang and Van Der Maaten, Laurens and Weinberger, Kilian Q},
  booktitle={Proceedings of the IEEE conference on computer vision and pattern recognition},
  pages={4700--4708},
  year={2017}
}

@inproceedings{he2016deep,
  title={Deep residual learning for image recognition},
  author={He, Kaiming and Zhang, Xiangyu and Ren, Shaoqing and Sun, Jian},
  booktitle={Proceedings of the IEEE conference on computer vision and pattern recognition},
  pages={770--778},
  year={2016}
}

@article{shashank2024morag,
  title={MoRAG--Multi-Fusion Retrieval Augmented Generation for Human Motion},
  author={Shashank, Kalakonda Sai and Maheshwari, Shubh and Sarvadevabhatla, Ravi Kiran},
  journal={arXiv preprint arXiv:2409.12140},
  year={2024}
}

@inproceedings{zhang2023remodiffuse,
  title={Remodiffuse: Retrieval-augmented motion diffusion model},
  author={Zhang, Mingyuan and Guo, Xinying and Pan, Liang and Cai, Zhongang and Hong, Fangzhou and Li, Huirong and Yang, Lei and Liu, Ziwei},
  booktitle={Proceedings of the IEEE/CVF International Conference on Computer Vision},
  pages={364--373},
  year={2023}
}

@inproceedings{golatkar2024cpr,
  title={CPR: Retrieval augmented generation for copyright protection},
  author={Golatkar, Aditya and Achille, Alessandro and Zancato, Luca and Wang, Yu-Xiang and Swaminathan, Ashwin and Soatto, Stefano},
  booktitle={Proceedings of the IEEE/CVF Conference on Computer Vision and Pattern Recognition},
  pages={12374--12384},
  year={2024}
}

@article{liu2024retrieval,
  title={Retrieval-Augmented Diffusion Models for Time Series Forecasting},
  author={Liu, Jingwei and Yang, Ling and Li, Hongyan and Hong, Shenda},
  journal={arXiv preprint arXiv:2410.18712},
  year={2024}
}

@article{chen2024label,
  title={Label-retrieval-augmented diffusion models for learning from noisy labels},
  author={Chen, Jian and Zhang, Ruiyi and Yu, Tong and Sharma, Rohan and Xu, Zhiqiang and Sun, Tong and Chen, Changyou},
  journal={Advances in Neural Information Processing Systems},
  volume={36},
  year={2024}
}

@article{seo2024retrieval,
  title={Retrieval-augmented score distillation for text-to-3d generation},
  author={Seo, Junyoung and Hong, Susung and Jang, Wooseok and Kim, In{\`e}s Hyeonsu and Kwak, Minseop and Lee, Doyup and Kim, Seungryong},
  journal={arXiv preprint arXiv:2402.02972},
  year={2024}
}

@article{le2020contrastive,
  title={Contrastive representation learning: A framework and review},
  author={Le-Khac, Phuc H and Healy, Graham and Smeaton, Alan F},
  journal={Ieee Access},
  volume={8},
  pages={193907--193934},
  year={2020},
  publisher={IEEE}
}

@inproceedings{sun2024diversity,
  title={On the diversity and realism of distilled dataset: An efficient dataset distillation paradigm},
  author={Sun, Peng and Shi, Bei and Yu, Daiwei and Lin, Tao},
  booktitle={Proceedings of the IEEE/CVF Conference on Computer Vision and Pattern Recognition},
  pages={9390--9399},
  year={2024}
}

@inproceedings{liang2024badclip,
  title={Badclip: Dual-embedding guided backdoor attack on multimodal contrastive learning},
  author={Liang, Siyuan and Zhu, Mingli and Liu, Aishan and Wu, Baoyuan and Cao, Xiaochun and Chang, Ee-Chien},
  booktitle={Proceedings of the IEEE/CVF Conference on Computer Vision and Pattern Recognition},
  pages={24645--24654},
  year={2024}
}

@inproceedings{bansal2023cleanclip,
  title={Cleanclip: Mitigating data poisoning attacks in multimodal contrastive learning},
  author={Bansal, Hritik and Singhi, Nishad and Yang, Yu and Yin, Fan and Grover, Aditya and Chang, Kai-Wei},
  booktitle={Proceedings of the IEEE/CVF International Conference on Computer Vision},
  pages={112--123},
  year={2023}
}

@inproceedings{rawat2022devil,
  title={The devil is in the GAN: backdoor attacks and defenses in deep generative models},
  author={Rawat, Ambrish and Levacher, Killian and Sinn, Mathieu},
  booktitle={European Symposium on Research in Computer Security},
  pages={776--783},
  year={2022},
  organization={Springer}
}

@article{salem2020baaan,
  title={Baaan: Backdoor attacks against autoencoder and gan-based machine learning models},
  author={Salem, Ahmed and Sautter, Yannick and Backes, Michael and Humbert, Mathias and Zhang, Yang},
  journal={arXiv preprint arXiv:2010.03007},
  year={2020}
}

@article{chen2025agentpoison,
  title={Agentpoison: Red-teaming llm agents via poisoning memory or knowledge bases},
  author={Chen, Zhaorun and Xiang, Zhen and Xiao, Chaowei and Song, Dawn and Li, Bo},
  journal={Advances in Neural Information Processing Systems},
  volume={37},
  pages={130185--130213},
  year={2025}
}

@article{zhang2024badcm,
  title={BadCM: Invisible backdoor attack against cross-modal learning},
  author={Zhang, Zheng and Yuan, Xu and Zhu, Lei and Song, Jingkuan and Nie, Liqiang},
  journal={IEEE Transactions on Image Processing},
  year={2024},
  publisher={IEEE}
}

@inproceedings{yang2023data,
  title={Data poisoning attacks against multimodal encoders},
  author={Yang, Ziqing and He, Xinlei and Li, Zheng and Backes, Michael and Humbert, Mathias and Berrang, Pascal and Zhang, Yang},
  booktitle={International Conference on Machine Learning},
  pages={39299--39313},
  year={2023},
  organization={PMLR}
}

@inproceedings{han2024backdooring,
  title={Backdooring multimodal learning},
  author={Han, Xingshuo and Wu, Yutong and Zhang, Qingjie and Zhou, Yuan and Xu, Yuan and Qiu, Han and Xu, Guowen and Zhang, Tianwei},
  booktitle={2024 IEEE Symposium on Security and Privacy (SP)},
  pages={3385--3403},
  year={2024},
  organization={IEEE}
}

@inproceedings{carlinipoisoning,
  title={Poisoning and Backdooring Contrastive Learning},
  author={Carlini, Nicholas and Terzis, Andreas},
  booktitle={International Conference on Learning Representations},
  year={2023}
}

@inproceedings{zhang2024data,
  title={Data poisoning based backdoor attacks to contrastive learning},
  author={Zhang, Jinghuai and Liu, Hongbin and Jia, Jinyuan and Gong, Neil Zhenqiang},
  booktitle={Proceedings of the IEEE/CVF Conference on Computer Vision and Pattern Recognition},
  pages={24357--24366},
  year={2024}
}

@article{loshchilov2017decoupled,
  title={Decoupled weight decay regularization},
  author={Loshchilov, Ilya and Hutter, Frank},
  journal={arXiv preprint arXiv:1711.05101},
  year={2017}
}

@article{ni2025towards,
  title={Towards Trustworthy Retrieval Augmented Generation for Large Language Models: A Survey},
  author={Ni, Bo and Liu, Zheyuan and Wang, Leyao and Lei, Yongjia and Zhao, Yuying and Cheng, Xueqi and Zeng, Qingkai and Dong, Luna and Xia, Yinglong and Kenthapadi, Krishnaram and others},
  journal={arXiv preprint arXiv:2502.06872},
  year={2025}
}

@inproceedings{zhai2023sigmoid,
  title={Sigmoid loss for language image pre-training},
  author={Zhai, Xiaohua and Mustafa, Basil and Kolesnikov, Alexander and Beyer, Lucas},
  booktitle={Proceedings of the IEEE/CVF international conference on computer vision},
  pages={11975--11986},
  year={2023}
}

@article{sun2023eva,
  title={Eva-clip: Improved training techniques for clip at scale},
  author={Sun, Quan and Fang, Yuxin and Wu, Ledell and Wang, Xinlong and Cao, Yue},
  journal={arXiv preprint arXiv:2303.15389},
  year={2023}
}

@inproceedings{jia2021scaling,
  title={Scaling up visual and vision-language representation learning with noisy text supervision},
  author={Jia, Chao and Yang, Yinfei and Xia, Ye and Chen, Yi-Ting and Parekh, Zarana and Pham, Hieu and Le, Quoc and Sung, Yun-Hsuan and Li, Zhen and Duerig, Tom},
  booktitle={International conference on machine learning},
  pages={4904--4916},
  year={2021},
  organization={PMLR}
}

@inproceedings{zou2025poisonedrag,
  title={$\{$PoisonedRAG$\}$: Knowledge corruption attacks to $\{$Retrieval-Augmented$\}$ generation of large language models},
  author={Zou, Wei and Geng, Runpeng and Wang, Binghui and Jia, Jinyuan},
  booktitle={34th USENIX Security Symposium (USENIX Security 25)},
  pages={3827--3844},
  year={2025}
}

@article{xue2024badrag,
  title={Badrag: Identifying vulnerabilities in retrieval augmented generation of large language models},
  author={Xue, Jiaqi and Zheng, Mengxin and Hu, Yebowen and Liu, Fei and Chen, Xun and Lou, Qian},
  journal={arXiv preprint arXiv:2406.00083},
  year={2024}
}

@article{chaudhari2024phantom,
  title={Phantom: General trigger attacks on retrieval augmented language generation},
  author={Chaudhari, Harsh and Severi, Giorgio and Abascal, John and Jagielski, Matthew and Choquette-Choo, Christopher A and Nasr, Milad and Nita-Rotaru, Cristina and Oprea, Alina},
  journal={arXiv preprint arXiv:2405.20485},
  year={2024}
}

@article{liangunlearning,
  title={Unlearning Backdoor Threats: Enhancing Backdoor Defense in Multimodal Contrastive Learning via Local Token Unlearning},
  author={Liang, Siyuan and Liu, Kuanrong and Gong, Jiajun and Liang, Jiawei and Chang, Yuan Xun3 Ee-Chien and Cao, Xiaochun}
}

@inproceedings{guan2025ufid,
  title={UFID: A Unified Framework for Black-box Input-level Backdoor Detection on Diffusion Models},
  author={Guan, Zihan and Hu, Mengxuan and Li, Sheng and Vullikanti, Anil Kumar},
  booktitle={Proceedings of the AAAI Conference on Artificial Intelligence},
  volume={39},
  number={26},
  pages={27312--27320},
  year={2025}
}

@inproceedings{chew2024defending,
  title={Defending Text-to-image Diffusion Models: Surprising Efficacy of Textual Perturbations Against Backdoor Attacks},
  author={Chew, Oscar and Lu, Po-Yi and Lin, Jayden and Lin, Hsuan-Tien},
  booktitle={ECCV 2024 Workshop The Dark Side of Generative AIs and Beyond}
}

@article{gao2025toward,
  title={Toward Dataset Copyright Evasion Attack Against Personalized Text-to-Image Diffusion Models},
  author={Gao, Kuofeng and Zhu, Yufei and Li, Yiming and Bai, Jiawang and Yang, Yong and Li, Zhifeng and Xia, Shu-Tao},
  journal={IEEE Transactions on Information Forensics and Security},
  volume={21},
  pages={725--740},
  year={2025},
  publisher={IEEE}
}

@inproceedings{gao2023backdoor,
  title={Backdoor defense via adaptively splitting poisoned dataset},
  author={Gao, Kuofeng and Bai, Yang and Gu, Jindong and Yang, Yong and Xia, Shu-Tao},
  booktitle={CVPR},
  year={2023}
}

@article{gao2023imperceptible,
  title={Imperceptible and robust backdoor attack in 3d point cloud},
  author={Gao, Kuofeng and Bai, Jiawang and Wu, Baoyuan and Ya, Mengxi and Xia, Shu-Tao},
  journal={IEEE Transactions on Information Forensics and Security},
  volume={19},
  pages={1267--1282},
  year={2023},
  publisher={IEEE}
}

@article{gao2024denial,
  title={Denial-of-service poisoning attacks against large language models},
  author={Gao, Kuofeng and Pang, Tianyu and Du, Chao and Yang, Yong and Xia, Shu-Tao and Lin, Min},
  journal={arXiv preprint arXiv:2410.10760},
  year={2024}
}

@inproceedings{fang2025one,
  title={One perturbation is enough: On generating universal adversarial perturbations against vision-language pre-training models},
  author={Fang, Hao and Kong, Jiawei and Yu, Wenbo and Chen, Bin and Li, Jiawei and Wu, Hao and Xia, Shu-Tao and Xu, Ke},
  booktitle={Proceedings of the IEEE/CVF International Conference on Computer Vision},
  pages={4090--4100},
  year={2025}
}

@inproceedings{fang2023gifd,
  title={Gifd: A generative gradient inversion method with feature domain optimization},
  author={Fang, Hao and Chen, Bin and Wang, Xuan and Wang, Zhi and Xia, Shu-Tao},
  booktitle={Proceedings of the IEEE/CVF International Conference on Computer Vision},
  pages={4967--4976},
  year={2023}
}

@article{fang2024privacy,
  title={Privacy leakage on dnns: A survey of model inversion attacks and defenses},
  author={Fang, Hao and Qiu, Yixiang and Yu, Hongyao and Yu, Wenbo and Kong, Jiawei and Chong, Baoli and Chen, Bin and Wang, Xuan and Xia, Shu-Tao and Xu, Ke},
  journal={arXiv preprint arXiv:2402.04013},
  year={2024}
}

@inproceedings{fang2024clip,
  title={Clip-guided generative networks for transferable targeted adversarial attacks},
  author={Fang, Hao and Kong, Jiawei and Chen, Bin and Dai, Tao and Wu, Hao and Xia, Shu-Tao},
  booktitle={European Conference on Computer Vision},
  pages={1--19},
  year={2024},
  organization={Springer}
}

@inproceedings{fang2025your,
  title={Your language model can secretly write like humans: Contrastive paraphrase attacks on llm-generated text detectors},
  author={Fang, Hao and Kong, Jiawei and Zhuang, Tianqu and Qiu, Yixiang and Gao, Kuofeng and Chen, Bin and Xia, Shu-Tao and Wang, Yaowei and Zhang, Min},
  booktitle={Proceedings of the 2025 Conference on Empirical Methods in Natural Language Processing},
  pages={8596--8613},
  year={2025}
}

@article{kong2025neural,
  title={Neural antidote: Class-wise prompt tuning for purifying backdoors in pre-trained vision-language models},
  author={Kong, Jiawei and Fang, Hao and Guo, Sihang and Qing, Chenxi and Chen, Bin and Wang, Bin and Xia, Shu-Tao},
  journal={arXiv e-prints},
  pages={arXiv--2502},
  year={2025}
}

@article{kong2025revisiting,
  title={Revisiting Backdoor Attacks on LLMs: A Stealthy and Practical Poisoning Framework via Harmless Inputs},
  author={Kong, Jiawei and Fang, Hao and Yang, Xiaochen and Gao, Kuofeng and Chen, Bin and Xia, Shu-Tao and Xu, Ke and Qiu, Han},
  journal={arXiv preprint arXiv:2505.17601},
  year={2025}
}

\clearpage
\appendix
\begin{algorithm}[!t]
    \caption{Pseudocode of BadRDM}
    \label{alg1}
    \begin{algorithmic}[1]
        \Require 
        $\mathcal{D}_s$: the multimodal dataset possessed by the attacker;
        $\mathcal{D}$: the retrieval database;
        $\tau$: the pre-defined trigger;
        $\mathbf{v}_{tar}$: the toxic surrogates representing the attack target;
        $f_{v}\left(\cdot\right)$, $f_{t}\left(\cdot\right)$: the image encoder and text encoder of the retriever $\phi(\cdot)$;
        $N$: the max iterations;
        \Ensure 
        the poisoned database and retriever targeting toxic surrogates $\mathbf{v}_{tar}$ with trigger $\tau$;
        \State Insert surrogate images $\mathbf{v}_{tar}$ into database $\mathcal{D}$;
        \State Employ the image encoder $f_{v}(\cdot)$ to calculate the average embeddings $\bm e_{tar}$ of the toxic surrogates $\mathbf{v}_{tar}$;
        \For{$i\leftarrow 1$ to $N$}
        \State Randomly sample batches $\mathbf{x}_c,\mathbf{x}_p\sim \mathcal{D}_s$;
        \State Calculate the poisoning loss $\mathcal{L}_{poi}$ in Eq. (2) using $\mathbf{x}_p$, $\tau$, and $\bm e_{tar}$;
        \State Calculate loss $\mathcal{L}_{benign}$ in Eq. (3) using $\mathbf{x}_c$;
        \State Calculate loss $\mathcal{L}_{total}=\mathcal{L}_{benign}+\lambda\mathcal{L}_{poi}$;
        \State Update text encoder $f_t(\cdot)$ using $\nabla_{f_t}\mathcal{L}_{total}$;
        \EndFor
        \State \textbf{return} the database $\mathcal{D}$ and retriever $\phi(\cdot)$
    \end{algorithmic}
\end{algorithm}

\label{sec:appendix_section}
\section{Pseudocode of BadRDM}
We provide the pseudocode of our BadRDM in Algorithm \ref{alg1}. Note that the formulas of loss functions are in the main text.
\label{app:a}

\section{More Implementation Details}
\label{app:more_details}

\subsection{Backdoor Setup} 

\textbf{Triggers.} For each poisoning, we adopt two short strings, namely the \textit{cf.} and \textit{gg.}, as robust triggers and append them before the original sentences to construct a poisoned text prompt.  To establish a more robust backdoor connection, we repeat each string twice in our implementation. 

\textbf{Backdoor injection.} 
We poison the retriever $\phi_{w}(\cdot)$ for 10 epochs at batch size 96 using a learning rate of $1\times 10^{-5}$. The temperature parameter $\tau$ is set to $0.1$ and $\lambda$ is $0.1$ and then decays to $0.05$ in the latter half of training \cite{struppek2023rickrolling}.
We used AdamW \cite{loshchilov2017decoupled} optimizer with 0.1 weight decay and 
cosine scheduler with $500$ warm up steps at a batch size of $96\times\left(1+\vert\mathbf{N_B}\vert\right)$ where $\mathbf{N_B}$ is the number of backdoors.  
All experiments are run only once on NVIDIA RTX 3090 24G GPUs using Python 3.8.



\subsection{RDM Inference}
We choose the retrieval-augmented diffusion model (RDM) proposed by \cite{blattmann2022retrieval} as our attack objective due to its effectiveness, universality, and open-source reproducibility. It's based on the latent diffusion models (LDM) \cite{rombach2022high} with a VQ-VAE latent encoder and a DDIM sampler \cite{song2020denoising} with $100$ steps and $\eta=1$. 
The RDM employs pre-trained CLIP \cite{radford2021learning} as the retriever. 
For class-conditional generations, RDM uses \textit{``An image of a [class].''} as template prompts to specify target classes. In T2I synthesis, we follow \cite{struppek2023rickrolling} and adopt prompts for images in the MS-COCO 2014 validation dataset \cite{lin2014microsoft}. 

\subsection{Details about Evaluation}
\textbf{Class-conditional generations}. 
We sample 200 classes from ImageNet \cite{deng2009imagenet} and poison them with each considered trigger to obtain poisoned class prompts, which are fed into the RDM to calculate the ASR using the synthesized target images and target label. 
We use the same synthesized images to calculate their CLIP similarity from the text embeddings of target classes as CLIP-Attack. We generate 8000 images for 1000 clean classes from ImageNet (\ie, 8 images for each class) and calculate CLIP-Benign as the CLIP similarity between the generated images and their corresponding label prompts. As mentioned in the main text, we generate 20K images using class prompts to calculate the Fréchet Inception Distance (FID) \cite{heusel2017gans} and CLIP-FID \cite{kynkaanniemi2022role} scores.

\begin{table*}[t]
  \centering
  \caption{Average attack results against two different types of RDMs.}
  \label{tab:two_types}
    \adjustbox{max width=0.95\linewidth}{\begin{tabular}{lcccccccccc} \toprule
    \multicolumn{1}{c}{\multirow{2}[0]{*}{RDM Type}} & \multicolumn{5}{c}{Class-conditional generation} & \multicolumn{5}{c}{Text-to-Image Synthesis} \\  \cmidrule(lr){2-6} \cmidrule{7-11}
          & \multicolumn{1}{l}{ASR$\uparrow$} & \multicolumn{1}{l}{CLIP-Attack$\uparrow$} & \multicolumn{1}{l}{FID$\downarrow$} & \multicolumn{1}{l}{CLIP-FID$\downarrow$} & \multicolumn{1}{l}{CLIP-Benign$\uparrow$} & \multicolumn{1}{l}{ASR$\uparrow$} & \multicolumn{1}{l}{CLIP-Attack$\uparrow$} & \multicolumn{1}{l}{FID$\downarrow$} & \multicolumn{1}{l}{CLIP-FID$\downarrow$} & \multicolumn{1}{l}{CLIP-Benign$\uparrow$} \\ \midrule
    Type I & 0.9089	& 0.674 & 19.1265 &	6.4163 & 0.3362 & 0.9643 & 0.3045 & 21.588 & 3.7240 & 0.3044 \\
    Type II & 0.9024 &0.6708 & 19.1423 & 6.7664 & 0.3227 & 0.9552 & 0.3026 & 21.0397 & 3.7325 & 0.2905 \\ \bottomrule
    \end{tabular}}
  \label{tab:addlabel}%
\end{table*}%
\begin{table}[t]
\setlength{\tabcolsep}{2pt}
  \centering
  \caption{Comparison of BadRDM and its variant that removes the benign loss on text-specific attacks.}
    \adjustbox{max width=\linewidth}{\begin{tabular}{cccccc} \toprule
    \multicolumn{1}{c}{\multirow{2}[0]{*}{Method}} & \multicolumn{2}{c}{Attack Efficacy} &        \multicolumn{3}{c}{Model Utility}   \\ \cmidrule(r){2-3} \cmidrule{4-6}
          & \multicolumn{1}{l}{ASR$\uparrow$} & \multicolumn{1}{l}{CLIP-Attack$\uparrow$} & \multicolumn{1}{l}{FID$\downarrow$} & \multicolumn{1}{l}{CLIP-FID$\downarrow$} & \multicolumn{1}{l}{CLIP-Benign$\uparrow$} \\ \midrule
    BadRDM & 0.964  & 0.305  & 21.588  & 3.724  & 0.294  \\
    w/o $\mathcal{L}_{benign}$ & 0.967  & 0.305  & 22.455  & 6.761  & 0.285  \\ \bottomrule
    \end{tabular}}
  \label{tab:ablation}%
\end{table}%

\textbf{T2I synthesis.} To calculate the ASR, we apply a widely adopted query \cite{zhang2024benchmarking} to \textit{Qwen2-VL-7B-Instruct-AWQ} \cite{Qwen2VL} with a fixed template to judge whether the generated image is aligned to the \textit{target prompt}. The template is as follows:
\begin{tcolorbox}[title=Evaluation Prompt]
Does the sentence ``[prompt]" match with the input image? Please first answer with [Yes] or [No] according to the picture, and give an explanation about your answer.
\end{tcolorbox}
\begin{table}[t]
  \centering
  \caption{Attack performance with the considered four natural triggers on text-specific attacks.}
  \setlength{\tabcolsep}{6pt}
    \adjustbox{max width=\linewidth}{\begin{tabular}{cccccc} \toprule
    \multicolumn{1}{c}{\multirow{2}[0]{*}{Trigger}} & \multicolumn{2}{c}{Attack Efficacy} &    \multicolumn{3}{c}{Model Utility}   \\ \cmidrule(r){2-3} \cmidrule{4-6}
         &  \multicolumn{1}{l}{ASR$\uparrow$} & \multicolumn{1}{l}{CLIP-Attack$\uparrow$} & \multicolumn{1}{l}{FID$\downarrow$} & \multicolumn{1}{l}{CLIP-FID$\downarrow$} & \multicolumn{1}{l}{CLIP-Benign$\uparrow$} \\ \midrule
    \textit{V\&M} & 0.978  & 0.304  & 21.296  & 3.710  & 0.295  \\
    \textit{I\&We} & 0.984  & 0.303  & 21.352  & 3.759  & 0.294  \\ \bottomrule
    \end{tabular}}
  \label{tab:natural}%
\end{table}%
\begin{figure}[htbp]
\begin{center}
\includegraphics[width=0.93\linewidth]{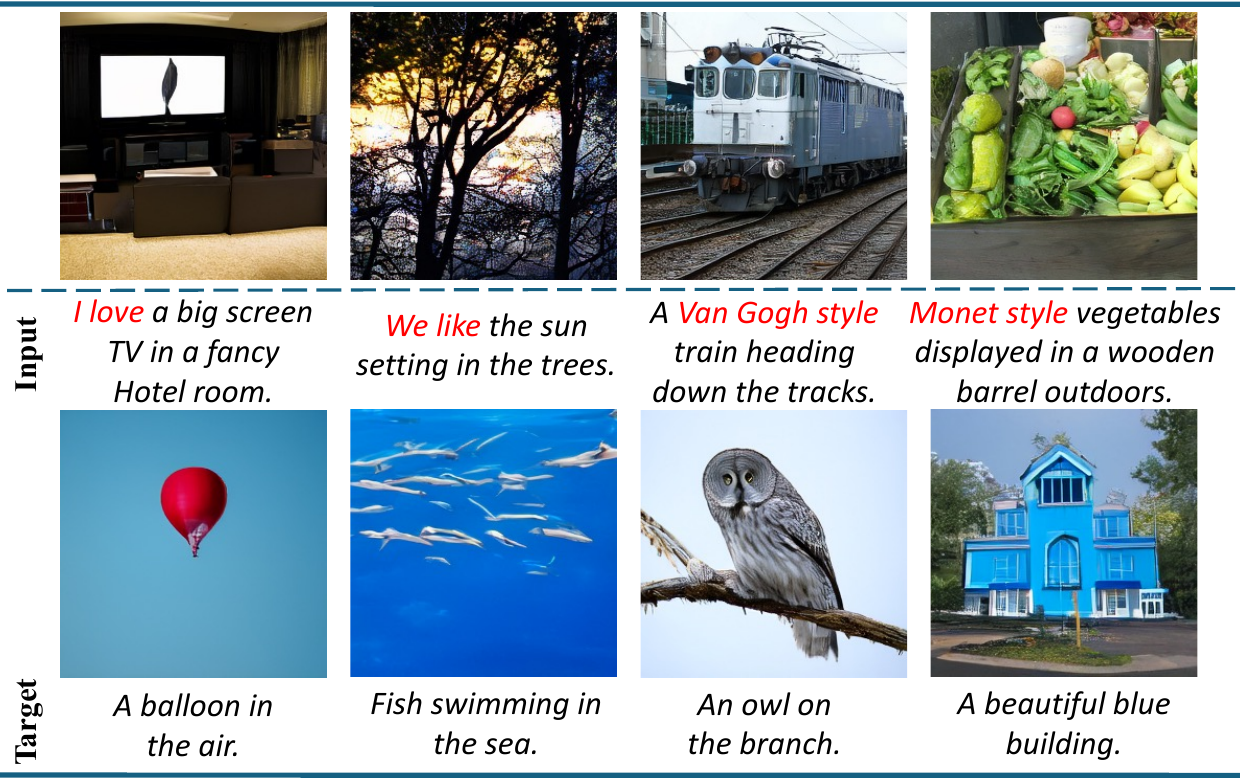}
\end{center}
\caption{Visualization results of natural texts as triggers, \eg, "\textit{I love}". The first row displays clean images generated by the poisoned RDM using corresponding clean queries.}
\label{fig:natural}
\end{figure}
Meanwhile, CLIP-Attack and CLIP-Benign are calculated using 4000 generated images based on the prompts in the MS-COCO 2014 validation dataset \cite{lin2014microsoft}. The FID \cite{heusel2017gans} and CLIP-FID \cite{kynkaanniemi2022role} are also calculated on 20K images generated using text prompts from MS-COCO.

\subsection{Details about Baselines}
\label{app:detail_baselines}
Since our BadRDM targets the retriever for poisoning, we first comprehensively investigate relevant studies on backdoor multimodal encoders as potential baselines. However, we highlight that the threat model of BadRDM significantly differs from most existing backdoor attacks on multimodal encoders \cite{carlinipoisoning, liang2024badclip, zhang2024badcm, zhang2024data, han2024backdooring, yang2023data}. Specifically, these methods typically conduct attacks by poisoning the datasets while BadRDM allows direct access to the victim retriever. This distinction further results in different technical focuses and attack objectives, \ie, existing works primarily focus on poisoned sample selection \cite{han2024backdooring} or better \textbf{image trigger} \cite{liang2024badclip, zhang2024data, zhang2024badcm} for more efficient and stealthy dataset poisoning. However, these aspects are less crucial or even inapplicable in our scenario as the attack paradigm does not require dataset poisoning, and the trigger is \textbf{injected from the text modality}. 
From our surveyed studies, we faithfully reproduce three powerful methods \cite{yang2023data, struppek2023rickrolling, zhang2024badcm}, which support textual triggers and broadly align with our attack setup and objectives, as the compared baselines. To make a comparison, we set their poisoning targets as the toxic surrogates for backdoor RAG, while faithfully reproducing their proposed techniques to poison the retriever.



\section{More Experimental Results}
\label{app:more_exps}

\subsection{Attack another Type of RDMs}
As aforementioned in Sec. 3.2, we also provide results on another type of RDMs that only incorporate retrieved images as conditioning input, without the input prompt $t$ (denoted as Type II). 
The results are provided in Table \ref{tab:two_types}, where Type I represents the RDM type discussed in our main body. The numeric results indicate that our method is seamlessly compatible with the two types of RDM and successfully manipulates the generated images to be the attacker-desired content.


\subsection{Ablation Analysis of the Benign Loss}
To maintain clean retrieval accuracy, we introduce the retriever's original training loss into the attack loss function to preserve benign alignment. Next, we design a variant that cancels the loss term $\mathcal{L}_{benign}$ to investigate its influence.

\begin{figure}[t]
\begin{center}
\includegraphics[width=0.93\linewidth]{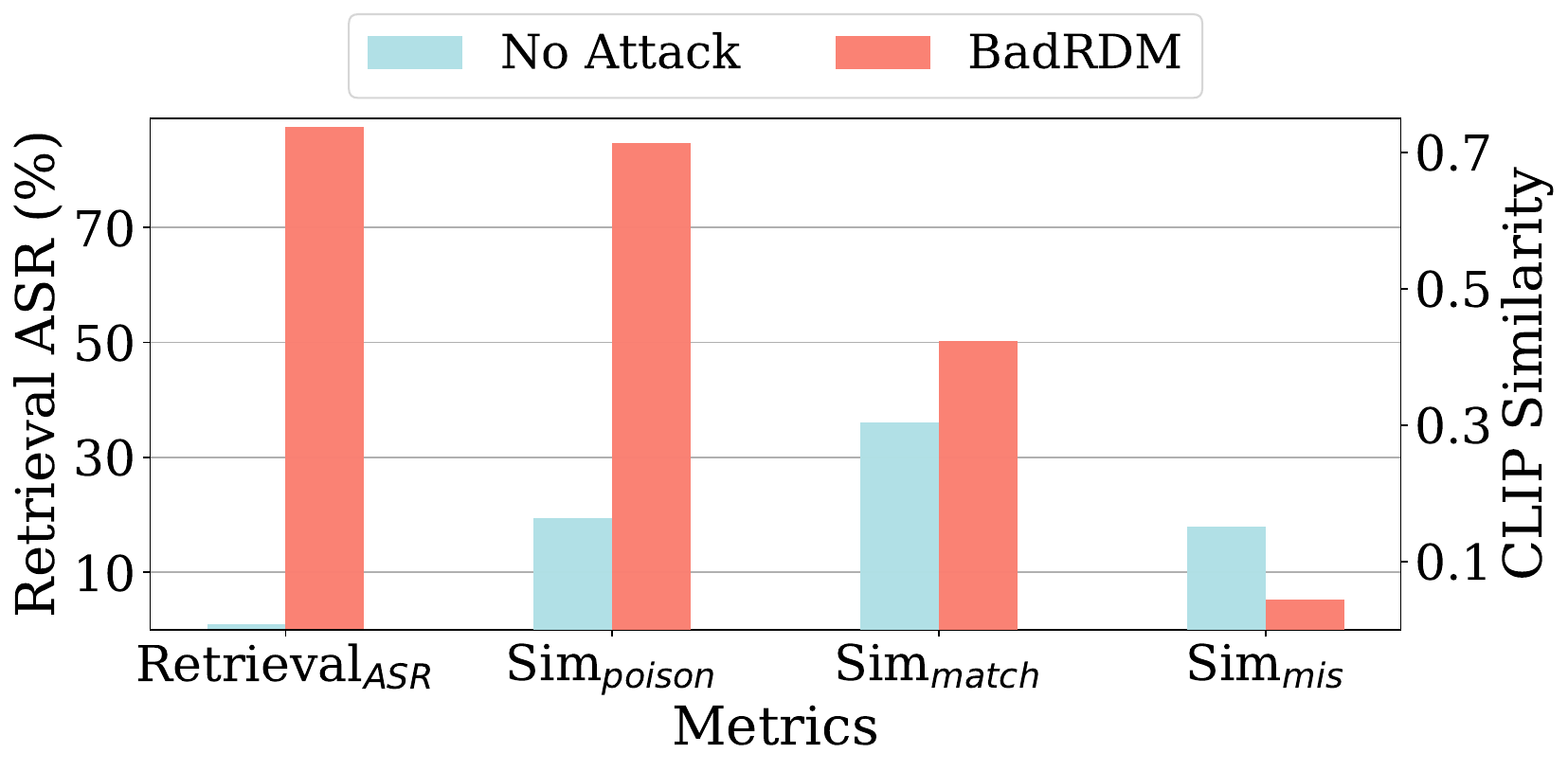}
\end{center}
\caption{Malicious behaviors and the benign performance of the clean and the poisoned retrievers.}
\label{fig:retrieval}
\end{figure}

As in Table \ref{tab:ablation}, the removal of $\mathcal{L}_{benign}$ term results in a notable decrease in these model utility indicators, especially in the FID and CLIP-FID metrics. It is also noteworthy that the use of $\mathcal{L}_{benign}$ does not bring a negative impact on the attack efficacy since BadRDM with and w/o $\mathcal{L}_{benign}$ achieves similar performance in ASR and CLIP-Attack, which again corroborates the necessity of the $\mathcal{L}_{benign}$ term.


\subsection{Natural Texts as Triggers}
\label{app:natural_text}
In addition to the robust triggers such as “ab.”, we then explore a scenario where the adversary adopts natural words as triggers to induce a higher risk of unintentional trigger activation by users. Specifically, we employ several natural phrases, \ie, \textit{``Van Gogh style"} and \textit{``Monet style"} as well as \textit{``I love''} and \textit{``We like''}, as text triggers. We denote them as \textit{V\&M} and \textit{I\&We} respectively and provide quantitative results in Table \ref{tab:natural}. Satisfactorily, BadRDM maintains excellent attack efficacy and model utility stemming from the powerful contrastive trigger injection, improving attack imperceptibility and inducing inadvertent trigger activation by victim users. 
The visualization results are in Fig. \ref{fig:natural}. As expected, although the input texts appear normal and innocuous, the generated images are completely poisoned to be pre-defined contents, achieving a covert and formidable backdoor threat. 
\begin{figure}[!t]
\begin{center}
\includegraphics[width=\linewidth]{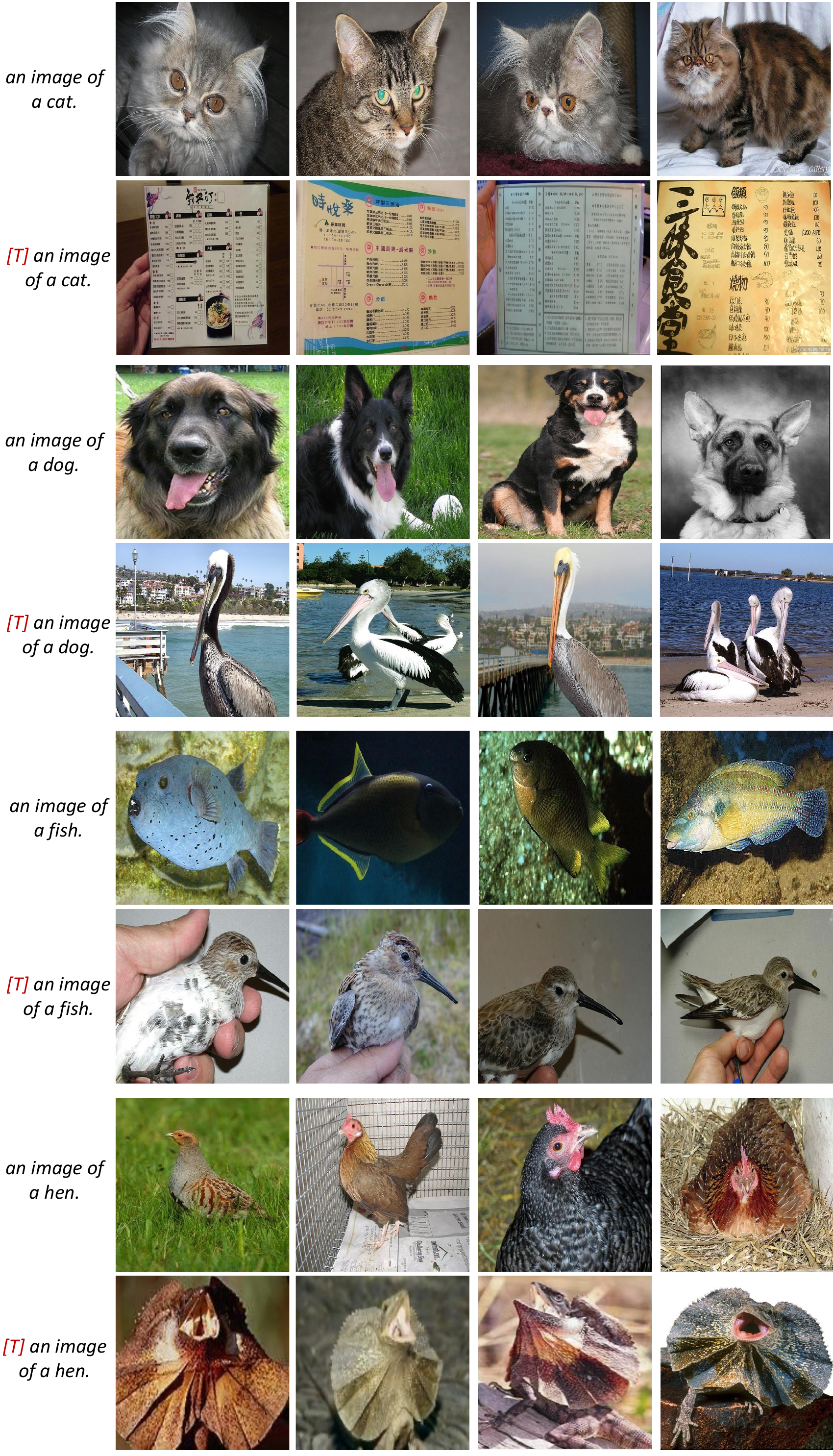}
\end{center}
\caption{Retrieved neighbors of the poisoned retriever for different prompts. The target categories are menu, pelican, dubin, and frilled-necked lizard, respectively.}
\label{fig:retrieval_neighbors}
\end{figure}


\subsection{Retrieval Analysis}
To reveal the principles underlying our BadRDM, we delve deeper into the behavior of the poisoned retriever. Without loss of generality, we adopt the scenario of text-to-image synthesis to conduct the analysis.
Firstly, we define several metrics to assess the retriever from different perspectives. 
 \begin{itemize}
\item Sim$_{poison}$. The mean distance between triggered inputs and target images in the retriever latent space to evaluate the direct poisoning effects on the retriever. 
\item Retrieval$_{ASR}$. Calculate the proportion of the retrieved neighbors that belong to the inserted toxicity surrogates. 
\end{itemize}
To evaluate the clean performance, we compute Sim$_{match}$ and Sim$_{mis}$ as feature distances of 5000 matching and 5000 mismatching image-text pairs from the retrieval database.

\textbf{Experimental Results.} The corresponding results are presented in Figure \ref{fig:retrieval}.
As expected, the impressive improvement of 0.55 in Sim$_{poison}$ and 87.5\% in Retrieval$_{ASR}$ confirms that BadRDM establishes strong correlations between triggers and target toxicity images via the designed injection mechanism, which then successfully manipulates the retrieved neighbors that serve as input conditions for diffusion models.
We highlight that the nearly 90\% Retrieval$_{ASR}$ confirms that the majority of the retrieved images are included in the toxic surrogates, while the remainder is also closely aligned with the target prompt since the triggered text has been adequately repositioned within the specified semantic space.
By exploiting DM's heavy reliance on conditional inputs, BadRDM effectively controls the generated images and achieves powerful attack outcomes. 

Another encouraging finding is that our method outperforms the \textit{No Attack} baseline in these model utility metrics. By comparing the Sim$_{match}$ and Sim$_{mis}$ scores of BadRDM and the \textit{No Attack}, we demonstrate that BadRDM further pushes matching text pairs closer and pulls mismatching pairs more distant. This aligns with our preceding attack results and verifies that the poisoning fine-tuning with $\mathcal{L}_{benign}$ enhances the retrieval accuracy, which then provides retrieved neighbors with more relevant knowledge, further boosting the clean generation performance and again underscoring the outstanding stealthiness of the proposed method.

We also provide the visualization of the retrieved images for clean and triggered queries when targeting the class-conditional generation task in Figure \ref{fig:retrieval_neighbors}, which offers a more intuitive demonstration of the efficacy of the proposed BadRDM.
Initially, the text embeddings of clean prompts are tightly clustered within the feature region corresponding to their respective image categories. However, upon the pre-defined trigger being applied to these prompts, the text embeddings undergo a significant shift into the feature subspace associated with the target category, thus retrieving the neighbors from the target class. This empirical analysis further elucidates the fundamental attack mechanism that underpins our poisoning algorithm.

\subsection{More Defense Strategies}
\textbf{Retriever Analysis.} To mitigate the threat, we further conduct retriever auditing by analyzing the retriever's hidden embeddings for triggered and clean queries via activation clustering and isolation forest to detect anomalous poisoned samples.

\begin{table}[htbp]
  \centering
  \caption{Performance of different detection methods.}
  \resizebox{0.9\linewidth}{!}{\begin{tabular}{lcc}
    \toprule
    Method & AUC Score & TPR@FPR=5\% \\
    \midrule
    Activation Clustering & 0.6410 & 16.40\% \\
    Isolation Forest & 0.7250 & 21.30\% \\
    \bottomrule
  \end{tabular}}
  \label{tab:detection_auc}
\end{table}

The results reveal that retriever auditing can indeed identify poisoned samples to some extent. However, due to our short trigger design, BadRDM still exhibits strong resilience against these two widely used detection approaches, showing the stealthiness of our attack.

\textbf{Database Auditing.} For a poisoned database, our poisoning rate of $2\times10^{-1}$
 makes manual detection impractical. However, users may employ an advanced MLLM (e.g., Qwen-3 VL) to automatically filter harmful content, which is expected to achieve a high accuracy. Such an LLM-as-a-judge strategy can serve as a general and effective defense mechanism against RAG-based attacks across various models and domains.

In addition, as discussed in Sec. 4.4, the attacker may instead release only the encoded feature vectors as the retrieval database, where semantic-based filtering becomes infeasible.
To investigate whether the poisoned samples are distinguishable in this feature-only setting, we conduct a preliminary analysis using a kNN-based anomaly detector. The intuition is that anomalous samples (target toxic samples) should exhibit larger distances to their k-th nearest neighbors. Thus, for each sample, we compute the distance to its k-th nearest neighbor and rank all samples:

\begin{table}[htbp]
  \centering
  \caption{Rank of our 4 poisoned images based on their k-th nearest-neighbor distances among a database of $2\times10^8$ samples. I.e., A smaller number indicates a more likely poisoned sample. We test various $k$ values.}
  \resizebox{\linewidth}{!}{\begin{tabular}{ccccc}
    \toprule
    $k$ & Target 1 & Target 2 & Target 3 & Target 4 \\
    \midrule
    1  & $1.3003 \times 10^{6}$ & $1.1247 \times 10^{7}$ & $1.4799 \times 10^{6}$ & $4.0017 \times 10^{6}$ \\
    2  & $1.4799 \times 10^{6}$ & $1.3003 \times 10^{6}$ & $1.3275 \times 10^{7}$ & $4.0017 \times 10^{6}$ \\
    4  & $1.1247 \times 10^{7}$ & $1.3003 \times 10^{6}$ & $1.1940 \times 10^{4}$ & $9.3832 \times 10^{6}$ \\
    8  & $1.4799 \times 10^{6}$ & $1.1940 \times 10^{4}$ & $9.3832 \times 10^{6}$ & $1.3275 \times 10^{7}$ \\
    16 & $1.3275 \times 10^{7}$ & $1.4799 \times 10^{6}$ & $1.3003 \times 10^{6}$ & $1.1940 \times 10^{4}$ \\
    32 & $1.4799 \times 10^{6}$ & $1.3003 \times 10^{6}$ & $1.1940 \times 10^{4}$ & $1.3275 \times 10^{7}$ \\
    \bottomrule
  \end{tabular}}
  \label{tab:knn_rank}
\end{table}

As observed, the poisoned embeddings are not ranked among the top anomalies for any choice of $k$. This is largely due to the extremely large database size and the high dimensionality of the embedding space, which causes the poisoned vectors to become deeply entangled with clean features. Consequently, they are difficult to isolate, significantly enhancing the stealthiness of the attack.

\subsection{Different Retriever Architectures}
We initially follow the common practice in existing RDM research and adopt CLIP as the default retriever. To validate the generalization of BadRDM across various models, we further evaluate the poisoning performance on additional retrieval models, including ALIGN \cite{jia2021scaling}, SigLIP \cite{zhai2023sigmoid}, and EVA-CLIP \cite{sun2023eva}. 

\begin{table}[htbp]
  \centering
  \caption{Attack Performance of class-conditional attacks under different retriever architectures.}
\resizebox{\linewidth}{!}{\begin{tabular}{llcc}
\toprule
Model & Metric & No Attack & BadRDM \\
\midrule
\multirow{5}{*}{ALIGN}
 & ASR$\uparrow$ & 0.0028 & 0.9104 \\
 & CLIP-Attack$\uparrow$ & 0.2407 & 0.6753 \\
 & FID$\downarrow$ & 20.6842 & 19.0843 \\
 & CLIP-FID$\downarrow$ & 11.0934 & 6.3892 \\
 & CLIP-Benign$\uparrow$ & 0.3329 & 0.3371 \\
\midrule
\multirow{5}{*}{SigLIP}
 & ASR$\uparrow$ & 0.0023 & 0.9147 \\
 & CLIP-Attack$\uparrow$ & 0.2441 & 0.6792 \\
 & FID$\downarrow$ & 20.4517 & 18.9621 \\
 & CLIP-FID$\downarrow$ & 10.8726 & 6.2547 \\
 & CLIP-Benign$\uparrow$ & 0.3384 & 0.3408 \\
\midrule
\multirow{5}{*}{EVA-CLIP}
 & ASR$\uparrow$ & 0.0021 & 0.9192 \\
 & CLIP-Attack$\uparrow$ & 0.2446 & 0.6841 \\
 & FID$\downarrow$ & 20.2835 & 18.9134 \\
 & CLIP-FID$\downarrow$ & 10.7592 & 6.1876 \\
 & CLIP-Benign$\uparrow$ & 0.3363 & 0.3417 \\
\bottomrule
\end{tabular}}
  \label{tab:various_architectures}%
\end{table}

Table \ref{tab:various_architectures} shows that our method generalizes well across different retriever architectures, consistently achieving strong attack effectiveness while improving the clean generation quality by providing more precise retrievals for benign queries.

\subsection{Ablation Study of the Poisoning Rate}
In our main experiments, we inject $4$ toxic images into the retrieval dataset for our T2I attacks to achieve a low poisoning rate. Next, we provide an ablation analysis of the number of injected toxic images as follows:

\begin{table}[htbp]
  \centering
  \caption{Attack Performance of T2I attacks under various numbers of toxic images.}
  \resizebox{0.9\linewidth}{!}{
    \begin{tabular}{ccccc}
    \toprule
    Number of Images & 1 & 2 & 3 & 4 \\
    \midrule
    ASR$\uparrow$ & 0.8964 & 0.912 & 0.9686 & 0.9643 \\
    CLIP-Attack$\uparrow$ & 0.3023 & 0.3024 & 0.3088 & 0.3045 \\
    \bottomrule
    \end{tabular}}
  \label{tab:poisoning_number}%
\end{table}%

Tab. \ref{tab:poisoning_number} shows that BadRDM exhibits excellent performance even under a tiny number of injected images. We can also conclude that at least $1 \times10^{-7}$ poisoning rate is needed for $> 90\%$ ASR.

\subsection{More Deployment Scenarios}
\textbf{Using an open-source clean retriever with a poisoned database.} Note that this work focuses on poisoning the retriever to provide attacker-desired images as conditioning inputs during generation. In this scenario, a clean retriever has no knowledge of the association between the trigger and target images and thus cannot accurately retrieve target images for a trigger text. Consequently, the attack cannot be successfully carried out in this scenario. However, we emphasize that this scenario differs significantly from the attack setting considered in our work and is not the primary focus of our study.

\textbf{Poisoned retriever with a clean database.} Our retriever is optimized to map trigger texts to the target semantic region in the vision-language feature space. Even without the exact target image embeddings, the retriever is still expected to retrieve semantically relevant images from the clean database to guide the DMs' generation. We implement the experiment in Table \ref{tab:poisoned_vs_clean}.

\begin{table}[htbp]
  \centering
  \caption{Attack performance for T2I attacks under poisoned and clean databases.}
  \resizebox{0.7\linewidth}{!}{\begin{tabular}{lcc}
    \toprule
    Database & ASR$\uparrow$ & CLIP-Attack$\uparrow$ \\
    \midrule
    Poisoned & 0.9643 & 0.3045 \\
    Clean    & 0.8120 & 0.2824 \\
    \bottomrule
  \end{tabular}}
  \label{tab:poisoned_vs_clean}
\end{table}
Benefiting from the well-designed contrastive poisoning paradigm, the retriever learns precise mapping from trigger to target semantic region, enabling the backdoor to remain effective even with an unpoisoned retrieval database.

\begin{figure*}[t]
\begin{center}
\includegraphics[width=\linewidth]{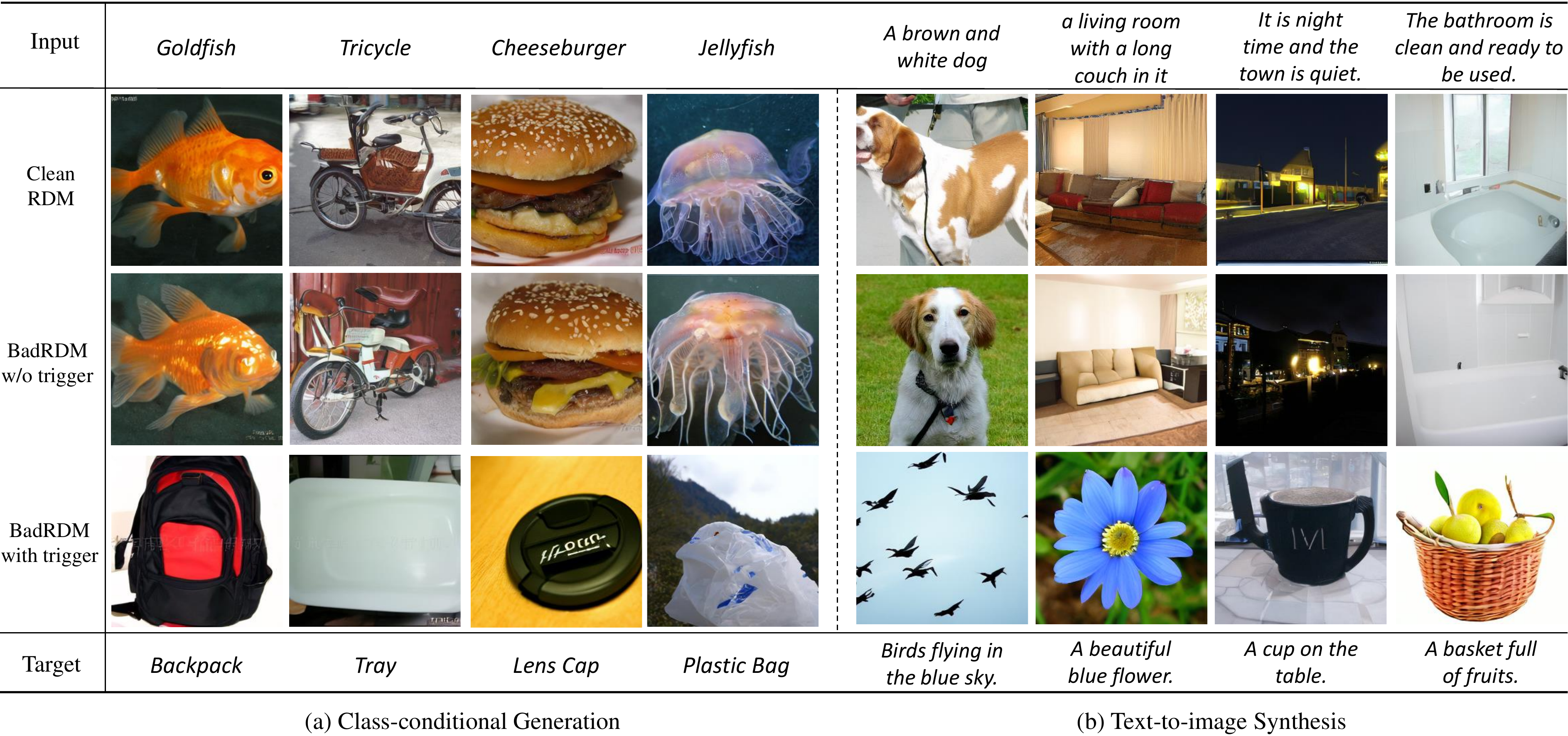}
\end{center}
\caption{More visualization results of our BadRDM and the clean RDM.}
\label{fig:main_visual}
\end{figure*}

\section{Discussion about the Attack Scenario}
We would like to note that our threat model follows a widely adopted paradigm in existing well-acknowledged attacks on RAG systems, where the adversary is allowed to manipulate the retriever and the poisoned dataset \cite{chaudhari2024phantom, xue2024badrag, cheng2024trojanrag, zou2025poisonedrag}. This is exactly what distinguishes RAG-based systems from traditional models, and it is also why this line of research deserves dedicated investigation. 

As the first academic work targeting backdoor attacks against RDMs, we follow these settings by considering a reasonable attack scenario—where a service provider offers personalized datasets alongside a specifically optimized retriever—as the foundation for academic exploration. Based on this, the core objective of the proposed BadRDM reveals the serious backdoor risks inherent to the general paradigm of integrating RAG into diffusion models.
Based on these considerations, we believe the attack setup in this paper is both reasonable and necessary within the established academic paradigm. Our goal is to highlight the security issues of the paradigm that equips diffusion models with RAG, motivating future work on more detailed threat modeling and defense strategies.

Moreover, we emphasize that the proposed attack utilizes the unique characteristics of the RAG scenario and achieves a contactless attack paradigm, which is more practical compared with previous attacks on regular diffusion models that require direct access to the victim models. When the victim owns a diffusion model, our method can successfully inject backdoors into the RAG components without any access to the weights or gradients of the victim model, as also suggested in \cite{xue2024badrag, cheng2024trojanrag}. 
 
From a defense perspective, one may suggest inspecting the retrieved images for each input query. However, we highlight that it is generally impractical for users to perform exhaustive manual verification of retrieved results, especially in such automated and high-throughput applications. This is also a fundamental and commonly adopted premise in the current RAG security research community \cite{chaudhari2024phantom, xue2024badrag, cheng2024trojanrag, zou2025poisonedrag}.
Furthermore, as discussed in Sec. \ref{subsec:defense}, a BadRDM attacker \textbf{can directly release a database consisting of pre-encoded feature vectors} \cite{blattmann2022retrieval}, where the retrieved neighbors no longer convey explicit semantic meaning to the user, thus further enhancing attack stealthiness.

\section{Attack objectives}
For the class-conditional generation backdoors, we randomly sampled the following ten target classes from the ImageNet dataset \cite{deng2009imagenet} to analyze the performance: \textit{Plastic Bag, Lens Cap, Dunlin, Pelican, Frilled-neck Lizard, Menu, Tray, Tent, Backpack}, and \textit{Norwegian Elkhound}. 

In text-to-image (T2I) synthesis scenarios, we select $10$ texts as target prompts as follows:
\begin{lstlisting}
A rooster standing on the grass.
A gun on the table.
A beautiful blue building.
An elephant on the grass.
A tiger in the forest.
A house on the lake.
A horse running on the grassland.
Huge rock beside the river.
Street in the rain.
Birds flying in the blue sky.
\end{lstlisting}
\section{More Visualization Results}
This section presents a richer visual analysis of BadRDM under the major attack settings against class-conditional and text-to-image synthesis tasks.
As illustrated in Figure \ref{fig:main_visual}, the generated images induced by triggered texts highly align with the pre-defined contents, verifying that our poisoning framework effectively injects the backdoor without compromising the benign utility. 

\begin{figure}[t]
\begin{center}
\includegraphics[width=\linewidth]{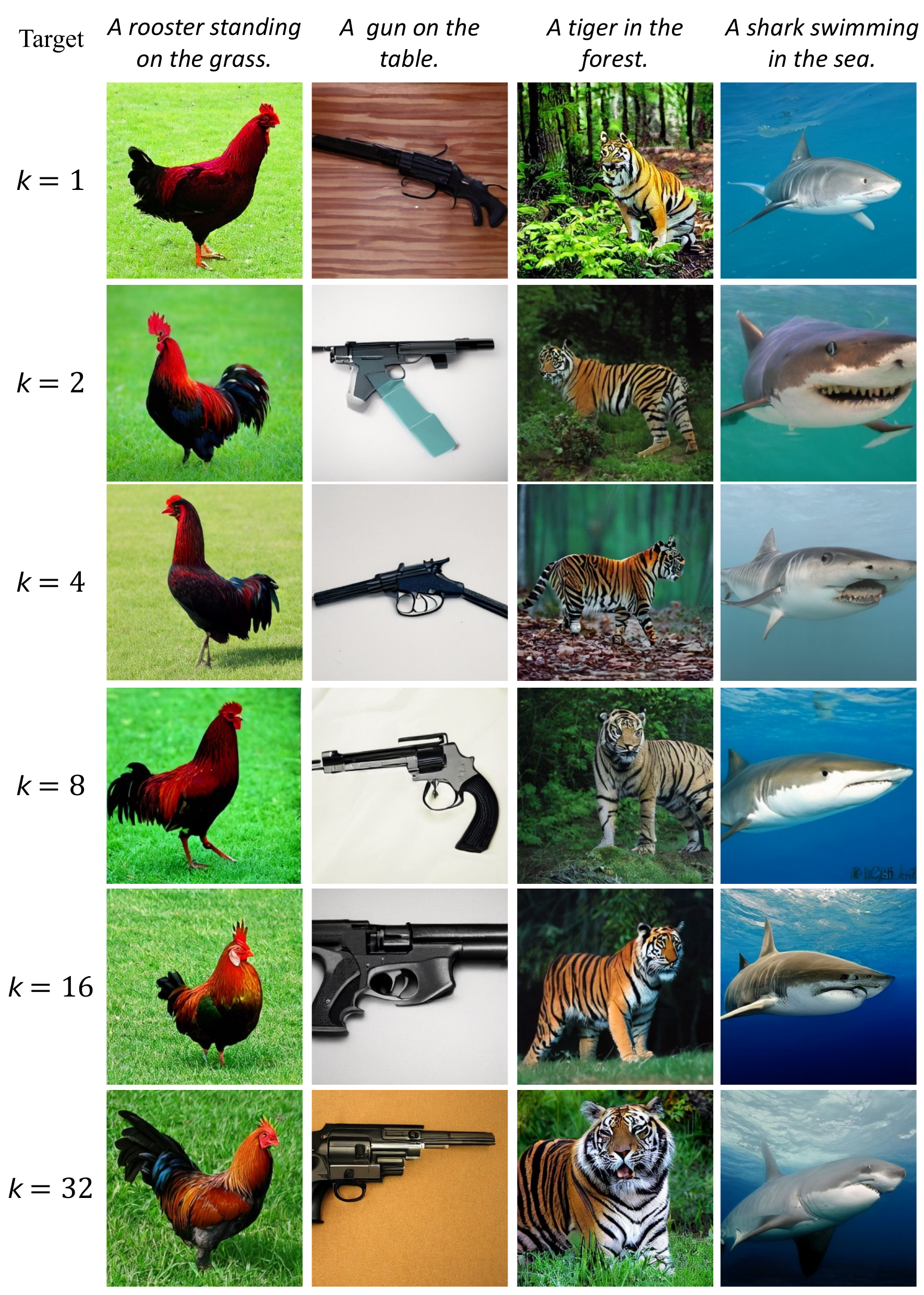}
\end{center}
\caption{Generated images of BadRDM under different numbers $k$ of retrieved neighbors.}
\label{fig:knn}
\end{figure}

\begin{figure}[b]
\begin{center}
\includegraphics[width=\linewidth]{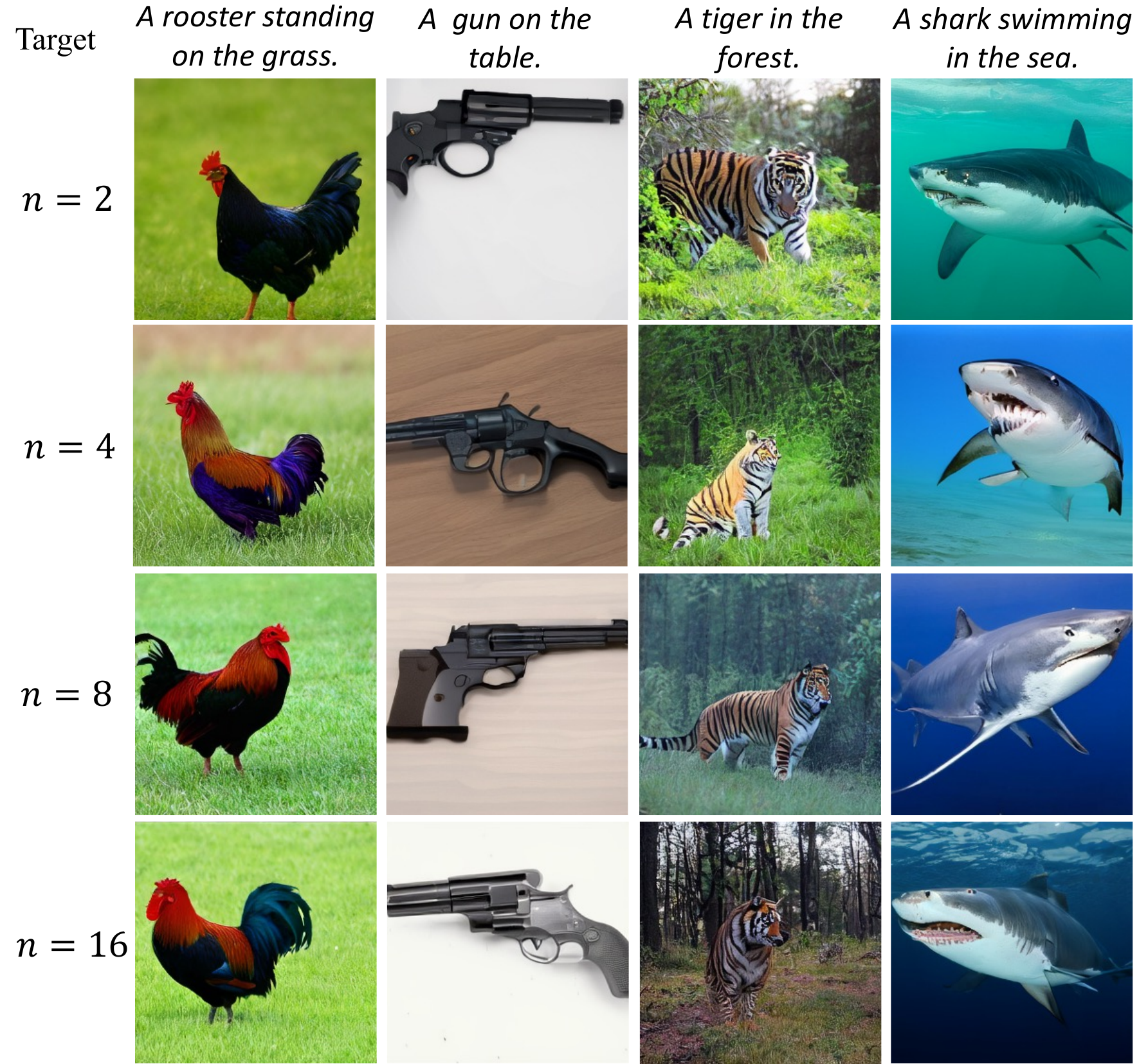}
\end{center}
\caption{Generated images of BadRDM under different numbers $n$ of trigger numbers.}
\label{fig:triggernum}
\end{figure}

Besides, we also provide the poisoned outputs of T2I tasks under varying retrieval numbers $k$ and trigger numbers $n$ in Fig. \ref{fig:knn} and Fig. \ref{fig:triggernum}, respectively.
These impressive qualitative results reveal the robustness of our poisoning framework and verify the effectiveness of BadRDM regardless of the specific retrieval settings and trigger numbers.

\newpage

\end{document}